\documentclass[twoside]{article}

\usepackage[accepted]{aistats2024_ArXiv}
\usepackage{amsmath}
\usepackage{amssymb}
\usepackage{mathtools}
\usepackage{amsthm}

\theoremstyle{plain}
\newtheorem{theorem}{Theorem}[section]
\newtheorem{proposition}[theorem]{Proposition}

\theoremstyle{definition}

\theoremstyle{remark}

\newtheorem{example}[theorem]{Example}

\usepackage{subcaption}
\usepackage{epsf}
\usepackage{fancyhdr}
\usepackage{graphics}
\usepackage{psfrag}

\usepackage{wrapfig}
\usepackage{float}

\usepackage{xr-hyper}
\usepackage{hyperref}

\usepackage{algorithmic}

\usepackage{color}

\usepackage{dsfont}
\usepackage{bm,bbm}
\usepackage{multirow}
\usepackage{balance}

\def\TW{\textsf{TW}}
\def\RT{\textsf{RT}}

\usepackage{verbatim}

\makeatletter
\def\munderbar#1{\underline{\sbox\tw@{$#1$}\dp\tw@\z@\box\tw@}}
\makeatother

\newcommand{\be}{\begin{equation}}
\newcommand{\ee}{\end{equation}}
\newcommand{\bea}{\begin{equation*}\begin{aligned}}
\newcommand{\eea}{\end{aligned}\end{equation*}}

\newcommand{\R}{\mathbb{R}}
\newcommand{\N}{\mathbb{N}}

\newcommand{\mbb}{\mathbb}

\newcommand{\calP}{{\mathcal P}}

\newcommand{\calT}{{\mathcal T}}
\newcommand{\calW}{{\mathcal W}}

\newcommand{\calR}{{\mathcal R}}
\newcommand{\calB}{{\mathcal B}}
\newcommand{\calU}{{\mathcal U}}

\newcommand{\UU}{\mbb U}

\newcommand{\RR}{\mbb R}

\newcommand{\dd}{\mathrm{d}}

\newcommand{\Let}{\triangleq}

\newcommand{\norm}[1]{\left\lVert#1\right\rVert}

\usepackage[round]{natbib}

\begin{document}

%

%






\runningauthor{Tam Le, Truyen Nguyen, Kenji Fukumizu}

\twocolumn[

\aistatstitle{Optimal Transport for Measures with Noisy Tree Metric}

\aistatsauthor{Tam Le$^{\dagger, \ddagger}$ \And Truyen Nguyen$^{\diamond}$ \And  Kenji Fukumizu$^{\dagger}$ }


\vspace{0.5mm}
\aistatsaddress{
The Institute of Statistical Mathematics$^{\dagger}$ \\
The University of Akron$^{\diamond}$ \\
RIKEN AIP$^{\ddagger}$} 

]

\begin{abstract}
We study optimal transport (OT) problem for probability measures supported on a tree metric space. It is known that such OT problem (i.e., tree-Wasserstein (TW)) admits a closed-form expression, but depends fundamentally on the underlying tree structure over supports of input measures. In practice, the given tree structure may be, however, perturbed due to noisy or adversarial measurements. To mitigate this issue, we follow the max-min robust OT approach which considers the maximal possible distances between two input measures over an uncertainty set of tree metrics. In general, this approach is hard to compute, even for measures supported in one-dimensional space, due to its non-convexity and non-smoothness which hinders its practical applications, especially for large-scale settings. In this work, we propose \emph{novel uncertainty sets of tree metrics} from the lens of edge deletion/addition which covers a diversity of tree structures in an elegant framework. Consequently, by building upon the proposed uncertainty sets, and leveraging the tree structure over supports, we show that the robust OT also admits a closed-form expression for a fast computation as its counterpart standard OT (i.e., TW). Furthermore, we demonstrate that the robust OT satisfies the metric property and is negative definite. We then exploit its negative definiteness to propose \emph{positive definite kernels} and test them in several simulations on various real-world datasets on document classification and topological data analysis.
\end{abstract}



\section{INTRODUCTION}
\label{sec:introduction}

Optimal transport (OT) has become a popular approach for comparing probability measures. OT provides a set of powerful tools that can be utilized in various research fields such as machine learning~\citep{peyre2019computational, nadjahi2019asymptotic, titouan2019optimal, bunne2019, bunne2022proximal, janati2020entropic, muzellec2020missing,  paty2020regularity, mukherjee2021outlier, altschuler2021averaging, fatras2021unbalanced,  scetbon2021low, si2021testing, le2021flow, le2021adversarial, liu2021lsmi, nguyen2021optimal, pmlr-v151-takezawa22a, fan2022complexity, hua2023curved, nguyen2023energy}, statistics~\citep{mena2019statistical, pmlr-v99-weed19a, liu2022entropy, nguyen2022many, nietert2022outlier, pmlr-v151-wang22f, pham2024scalable}, or computer vision/graphics~\citep{rabin2011wasserstein, solomon2015convolutional, lavenant2018dynamical, nguyen2021point, Saleh_2022_CVPR}.


Following the recent line of research on leveraging tree structure to scale up OT problems~\citep{LYFC, sato2020fast, le2021ept, pmlr-v151-takezawa22a, yamada2022approximating}, in this work, we study OT problem for probability measures supported on a tree metric space. Such OT problem (i.e., tree-Wasserstein (TW)) not only admits a closed-form expression, generalizes the sliced Wasserstein (SW)\footnote{SW projects supports into a one-dimensional space and exploits the closed-form expression of the univariate OT.}~\citep{rabin2011wasserstein} (i.e., a tree is a chain), but also alleviates the limited capacity issue of SW to capture topological structure of input measures, especially in high-dimensional spaces, since it provides more flexibility and degrees of freedom by choosing a tree rather than a line~\citep{LYFC}. However, it depends fundamentally on the underlying tree structure over supports of input measures. Nevertheless, in practical applications, the given tree structure may be perturbed due to noisy or adversarial measurements. For examples, (i) edge lengths may be noisy; (ii) for a physical tree, node positions may be perturbed or under adversarial attacks; (iii) some connecting nodes may be merged into each other; or (iv) some nodes may be duplicated and their corresponding edge lengths are positive. 

For OT problem with noisy ground cost, a common approach in the literature is to consider the maximal possible distance between two input measures over an uncertainty set of ground metrics, i.e., the \emph{max-min robust OT}~\citep{pmlr-v97-paty19a, deshpande2019max, lin2020projection}. However, such approach usually leads to optimization problems which are challenging to compute due to their non-convexity and non-smoothness~\citep{pmlr-v97-paty19a, lin2020projection}, even for input measures supported in one-dimensional spaces~\citep{deshpande2019max}). Another approach instead considers the \emph{min-max robust OT} which is a convexified relaxation and is an upper bound of the max-min robust OT~\citep{alvarez2018structured, pmlr-v84-genevay18a, pmlr-v97-paty19a, dhouib2020swiss}. 

Various advantages of the max-min/min-max robust OT have been reported in the literature. For examples, (i) it reduces the sample complexity~\citep{pmlr-v97-paty19a, deshpande2019max}; (ii) it increases the robustness to noise~\citep{pmlr-v97-paty19a, dhouib2020swiss}; (iii) it helps to induce prior structure, e.g., to encourage mapping of subspaces to subspaces used for domain adaptation where it is desirable to transport samples in the same class together~\citep{alvarez2018structured}; and (iv) it improves the generated images for generative model with the Sinkhorn divergence loss (i.e., entropic regularized OT) since the default Euclidean ground metric for Sinkhorn divergence loss tends to generate images which are basically a blur of similar images~\citep{pmlr-v84-genevay18a}.

The robust OT approaches can be interpreted in light of robust optimization~\citep{ben2009robust, bertsimas2011theory} where there are uncertainty parameters, especially  when the uncertainty parameters are not stochastic. The robust optimization has many roots and precursors in the applied sciences, particularly in robust control (e.g., to address the problem of stability margin~\citep{keel1988robust}); in machine learning (e.g., maximum margin principal in support vector machines (SVM)~\citep{xu2009robustness}), in reinforcement learning (e.g., to alleviate the gap between simulation environment and corresponding real-world environment~\citep{morimoto2001robust, NEURIPS2022_robustRL}). It is also known that robust optimization has a close connection with regularization~\citep{el1997robust, xu2008robust, xu2009robustness, bertsimas2011theory}. More precisely, solutions of several regularized problems are indeed solutions to a non-regularized robust optimization problem, e.g., Tikhonov-regularized regression~\citep{el1997robust}, Lasso~\citep{xu2008robust}, or norm-regularized SVM~\citep{xu2009robustness}.

Another interpretation of the robust OT is given under the perspective of the game theory. To see this, consider two players: the first player (the minimizer) aims at aligning the two measures by choosing a transport plan between two input measures; and the second player (the adversary) resists to it by choosing ground metric from the set of admissible ground metrics~\citep{alvarez2018structured}. Therefore, the robust OT approach can also be interpreted as to provide a safe choice of transportation plan under noisy ground metric for OT problem.

We emphasize both max-min and min-max robust OT approaches have their own advantages for the OT problem with noisy ground metric. In this work, we focus on the max-min robust OT for measures with a noisy tree metric ground cost.\footnote{One can sample tree metric for measures with supports in Euclidean space (see~\citep{LYFC}) to leverage TW for general applications, especially large-scale setting.} At a high level, our main contributions are three-fold as follows:
\begin{itemize}
    \item (i) We propose novel uncertainty sets of tree metrics from the lens of edge deletion/addition which cover a diversity of tree structures in an elegant framework. Consequently, by building upon the proposed uncertainty sets, and leveraging the tree structure over supports, we derive closed-form expressions for the max-min robust OT for measures with noisy tree metric, which is fast for computation and scalable for large-scale applications.

    \item (ii) We show that the max-min robust OT for measures with noisy tree metric satisfies metric property and is negative definite. Accordingly, we further propose positive definite kernels\footnote{A review on kernels is given in \S A.1 (supplementary).} built upon the robust OT, which are required in many kernel-dependent machine learning frameworks.
    
    \item (iii) We empirically illustrate that the max-min robust OT for measures noisy tree metric is fast for computation with the closed-form expression. Additionally, the proposed robust OT kernels improve performances of the counterpart standard OT (i.e., TW) kernel in several simulations on various real-world datasets on document classification and topological data analysis (TDA) for measures with noisy tree metric.

\end{itemize}

    
The paper is organized as follows: we give a brief recap of OT with tree metric cost in \S\ref{sec:reminders}. In \S\ref{sec:RobustTW}, we propose novel uncertainty sets of tree metrics, and leverage them to derive a closed-form expression for the max-min robust OT for measures with noisy tree metric. We show that it satisfies metric property and is negative definite. Consequently, we propose positive definite kernels built upon the robust OT. In \S\ref{sec:related_work}, we discuss related work. In \S\ref{sec:experiments}, we evaluate the proposed robust OT kernels for measures with noisy tree metric on document classification and TDA, and conclude our work in \S\ref{sec:conclusion}. Detailed proofs for our theoretical results are placed in the supplementary (\S B). Additionally, we have released code for our proposed approach.\footnote{\url{https://github.com/lttam/RobustOT-NoisyTreeMetric}}

\textbf{Notations.} We write $\mathds{1}$ for the vector of ones, and use  $|E|$ to denote the cardinality of set $E$. For $1\leq p \leq  \infty$, its conjugate is denoted by $p'$, i.e., $p' \in [1,\infty]$ s.t. $\frac{1}{p} +\frac{1}{p'}=1$. In particular, $p'=\infty$ when $p=1$, and $p'=1$ when $p=\infty$. Let $\norm{\cdot}_p$ represent the  $\ell_p$-norm in $\R^{|E|}$, and 
 $\overline\calB_p(v, \lambda) \Let \{u \in \R^{|E|}: \|u - v\|_p \leq  \lambda\}$ be the closed $\ell_p$-ball centering at $v\in \R^{|E|}$ and with radius $\lambda > 0$. We denote $\delta_x$ as the Dirac function at $x$.

\section{A REVIEW OF OPTIMAL TRANSPORT WITH TREE METRIC COST}
\label{sec:reminders}

In this section, we give a brief recap of OT with tree metric cost. We refer the readers to~\citet{LYFC} and the supplementary (\S A.2--\S A.3) for further details.

\paragraph{Tree metric.} Let $\calT = (V, E)$ be a tree rooted at $r$ with nonnegative weights $\{w_e\}_{e \in E}$ (i.e., edge length), where $V$ and $E$ are the sets of vertices and edges respectively. For any two nodes $x, z\in V$, we write $[x, z]$ for the unique path on $\calT$ connecting $x$ and $z$. For an edge $e$, $\gamma_e$ denotes the set of all nodes $x$ such that the path $ [r, x]$ contains the edge $e$. That is,
\begin{equation}\label{gamma_e}
    \gamma_e \Let\left\{x \in V \mid e \subset [r, x] \right\}.
\end{equation}
Let $d_{\calT}$ be the tree metric on $\calT$, that is  $d_{\calT}: V \times V \to [0,\infty)$ with $d_{\calT}(x,z)$ equaling to the length of the path $[x, z]$. We denote $\calP(V)$ for the set of all Borel probability measures on the set of nodes $V$, and use $w = (w_e)_{e\in E} \in \R^{|E|}_{+}$ to denote the vector of edge lengths for the tree $\calT$. 

\paragraph{Optimal transport (OT).} For probability measures $\mu, \, \nu \in \calP(V)$, let $\calR(\mu, \nu)$ be the set of measures $\pi$ on the product space $V \times V$ such that $\pi(A \times V) = \mu(A)$ and $\pi(V \times B)=\nu(B)$ for all Borel sets $A, B \subset V$. By using tree metric $d_{\calT}$ as the ground cost, the $1$-Wasserstein distance $\calW_{\calT}$ between $\mu$ and $\nu$ is defined as follows:
\begin{equation}\label{W_1}
\calW_{\calT}(\mu, \nu) \Let \inf_{\pi \in \calR(\mu, \nu)} \int_{V \times V} \hspace{-1em} d_{\calT}(x, z) \pi(\dd x, \dd z). 
\end{equation}
In Problem \eqref{W_1}, the OT distance for measures with tree metric ground cost (i.e., tree-Wasserstein (TW)) depends fundamentally on the tree structure $\calT$, which is determined by (i) the vector of edge lengths, i.e, $w = (w_e)_{e\in E}$ for tree $\calT$; and (ii) supports of input measures on tree, i.e., corresponding nodes in tree $\calT$. Therefore, for noisy tree metric, it may cause harm to OT performances. To mitigate this issue, in this work, we follow the max-min robust OT approach which seeks the maximal possible distance between input measures over an uncertainty set of tree metrics.

\section{ROBUST OPTIMAL TRANSPORT FOR MEASURES WITH NOISY TREE METRIC}\label{sec:RobustTW}

In this section, we describe the max-min robust OT approach for measures with noisy tree metric. We propose novel uncertainty sets of tree metrics which play the fundamental role to derive closed-form expressions for the robust OT. We also show that the robust OT satisfies metric property and is negative definite. Consequently, we prove positive definite kernels built upon the robust OT for input probability measures.

\paragraph{Max-min robust OT.}  Let $\UU(\calT)$ denote a family of tree metrics for the given tree $\calT$. By considering $\UU(\calT)$ as the uncertainty set of tree metrics, the max-min robust OT between two input probability measures $\mu, \nu \in \calP(V)$ is defined as follows:
\begin{equation}\label{eq:maxmin_robustTW}
\hspace{-0.2em} \RT(\mu, \nu) \Let \max_{\hat \calT \in \UU(\calT)}  \min_{\pi \in \calR(\mu, \nu)}  \int_{V \times V} \hspace{-1em} d_{\hat \calT}(x, z) \pi(\dd x, \dd z).
\end{equation}
Due to the tree nature, e.g., discrete structure, hierarchical relations among tree nodes, it is challenging to construct an uncertainty set of tree metrics which not only covers trees with various edge lengths, but also a diversity of tree structures in an elegant framework for robust OT.

To overcome the challenge on tree structures, inspired by the tree edit distance~\citep{tai1979tree} which utilizes a sequence of operations to transform a tree structure into another, we propose novel uncertainty sets of tree metrics from the lens of edge deletion/addition which is capable to cover a diversity of tree structures. These uncertainty sets play a cornerstone to scale up the max-min robust OT for measures with noisy tree metric in Problem~\eqref{eq:maxmin_robustTW}.


\begin{figure*}[h]
  \vspace{-4pt}
  \begin{center}
\includegraphics[width=0.9\textwidth]{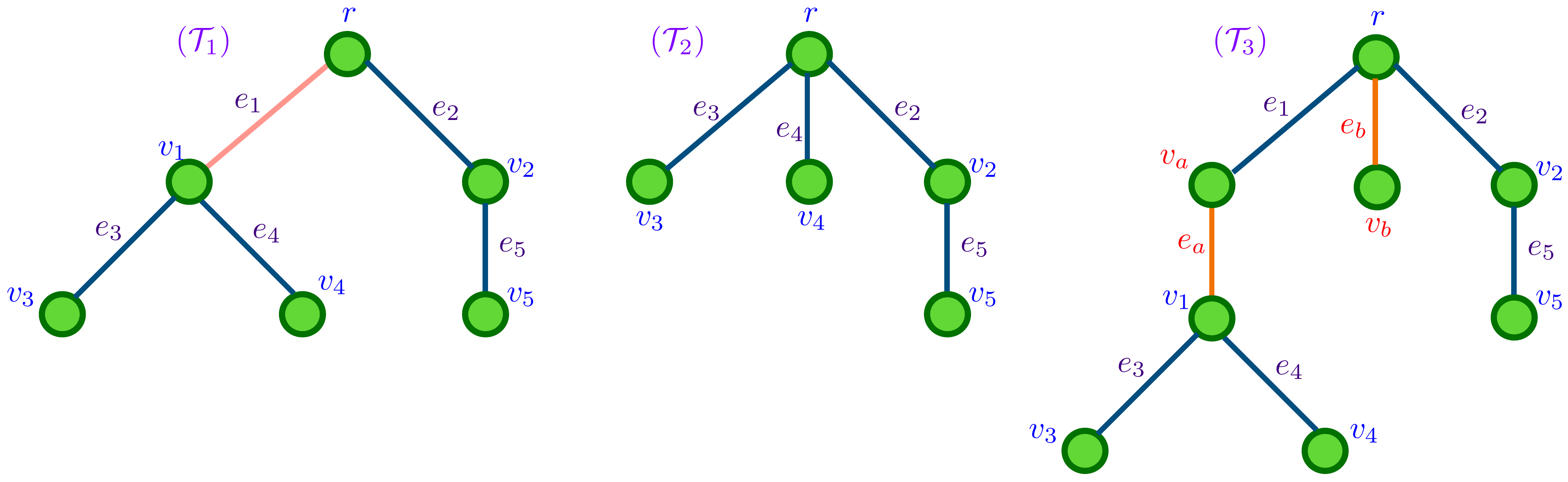}
  \end{center}
  \vspace{-6pt}
  \caption{An illustration of transforming a tree structure to another under the lens of edge addition/deletion. Given the binary tree structure $\calT_1$, if we collapse edge $e_1$ by merging vertex $v_1$ into the root vertex $r$ in $\calT_1$, we obtain the ternary tree structure $\calT_2$. Additionally, in tree $\calT_1$, if we duplicate vertex $v_1$ into $\{v_a, v_1 \}$, connect these two nodes by edge $e_a$; and add the vertex $v_b$ with edge $e_b$ between $v_b$ and root $r$, then we obtain the ternary tree structure $\calT_3$.}
  \label{fg:tree_transform}
\vspace{-4pt}
\end{figure*}

\paragraph{Uncertainty sets of tree metrics.} Our key observation is that the computation of OT between input measures with tree metric does not depend on edges which have $0$-length (i.e., $w_e=0$). We formally summarize it in Theorem~\ref{thrm:zeroedges}. 
\begin{theorem}\label{thrm:zeroedges}
Given tree $\calT$, denote $V$ as the set of vertices of $\calT$. Let $\calT'$ be a tree constructed from $\calT$ by collapsing its $0$-length edge, i.e., merging two corresponding vertices for an edge $e$ in $\calT$ with $w_e = 0$. Consequently, for any measure $\mu \in \calP(V)$ in tree $\calT$, its corresponding measure $\mu_{\calT'}$ in $\calT'$ is the same as the original measure $\mu$ in $\calT$, but mass of the supports in the collapsed edge $e$ is also merged for $\mu_{\calT'}$ in $\calT'$. Then, for any measures $\mu, \nu \in \calP(V)$ in $\calT$, we have
\begin{equation}
\calW_{\calT}(\mu, \nu) = \calW_{\calT'}(\mu_{\calT'}, \nu_{\calT'}).
\end{equation}
To simplify the notations, we also write $\calW_{\calT'}(\mu, \nu)$ for $\calW_{\calT'}(\mu_{\calT'}, \nu_{\calT'})$.
\end{theorem}
We give an illustration of transforming a tree structure into another under the lens of edge addition/deletion in Figure~\ref{fg:tree_transform}. To ease the understanding, let consider two following examples:

\begin{example}[Edge deletion for tree metric]\label{ex:edge_deletion} 
Let consider tree $\calT_1$ as in Figure~\ref{fg:tree_transform} and set $w_{e_1} = 0$. Tree $\calT_2$ is constructed from $\calT_1$ as in Figure~\ref{fg:tree_transform}. Given two probability measures $\mu = a_1\delta_{r} + a_2\delta_{v_1} + a_3\delta_{v_5}$ and $\nu = b_1\delta_{v_2} + b_2\delta_{v_4}$ in $\calT_1$ (i.e., $a_1, a_2, a_3, b_1, b_2 \ge 0; a_1+a_2+a_3=1; b_1 + b_2=1$), their corresponding measures in $\calT_2$ are $\mu_{\calT_2} = (a_1+a_2)\delta_{r} + a_3\delta_{v_5}$ and $\nu_{\calT_2} = \nu$ respectively. Then, we have 
\[
\calW_{\calT_1}(\mu, \nu) = \calW_{\calT_2}(\mu_{\calT_2}, \nu_{\calT_2}).
\]
\end{example}

\begin{example}[Edge addition for tree metric]\label{ex:edge_addition} 
Let consider tree $\calT_1$ as in Figure~\ref{fg:tree_transform}. Tree $\calT_3$ is constructed from $\calT_1$ as in Figure~\ref{fg:tree_transform}, and we set $w_{e_a} = w_{e_b} = 0$. Given two probability measures $\mu = a_1 \delta_{v_1} + a_2\delta_{v_5}$ and $\nu = b_1\delta_{v_2} + b_2\delta_{v_4}$ in $\calT_1$ (i.e., $a_1, a_2, b_1, b_2 \ge 0; a_1+a_2=1; b_1 + b_2=1$), their corresponding measures in $\calT_3$ are $\mu_{\calT_3} = a_{11} \delta_{v_a} + a_{12} \delta_{v_1} + a_2\delta_{v_5}$ where $a_{11}, a_{12} \ge 0; a_{11} + a_{12} = a_1$ and $\nu_{\calT_3} = \nu$ respectively. Then, regardless any split mass ($a_{11}, a_{12}$) in $\mu_{\calT_3}$ in $\calT_3$ from original $\mu$ in $\calT$, we have 
\[
\calW_{\calT_1}(\mu, \nu) = \calW_{\calT_3}(\mu_{\calT_3}, \nu_{\calT_3}).
\]
\end{example}

Therefore, we can add or delete edges with $0$-length on the given tree $\calT$ without changing the OT distance. More concretely, 
\begin{itemize}
\item (i) for edge deletion, we collapse edges with $0$-length in $\calT$ by merging the two corresponding vertices of those edges together (see Example~\ref{ex:edge_deletion}). 
\item (ii) for edge addition, we duplicate any vertex in $\calT$ and connect them with $0$-length edge (see Example~\ref{ex:edge_addition}). 
\end{itemize}
These actions help to transform the given tree structure $\calT$ into various tree structures, which play the fundamental role to construct uncertainty sets with diverse tree structures. Additionally, we further vary edge lengths of these tree structures to derive our novel uncertainty sets of tree metrics. 

The proposed uncertainty sets not only include tree metrics with a variety of tree structures (i.e., all subtree structures of the given tree $\calT$), but also tree metrics with varying edge lengths. In particular, one can further expand the expressiveness of these sets to cover more diverse tree structures by adding more edges with $0$-length for tree $\calT$ before varying its edge lengths (e.g., expanding tree $\calT_1$ into tree $\calT_3$ as in Figure~\ref{fg:tree_transform}, and using $\calT_3$ as the given tree), but it comes with a trade-off about the computation of the robust OT. 

More precisely, given tree $\calT = (V, E)$ with nonnegative weights $\{w_e\}_{e \in E}$, following Theorem~\ref{thrm:zeroedges}, it suffices to consider a family of tree metrics for $\calT$ where these tree structures share the same set of nodes $V$, the same root $r$, and the same set of edges $E$ as in $\calT$, but their edge lengths (i.e., edge weights) can be varied. To display this dependence on the vector of edge lengths, we will write $\calT(\hat w)$ for the tree in this family corresponding to the vector of edge lengths $\hat w = (\hat w_e)_{e\in E} \in \R^{|E|}_{+}$. In particular, we consider two approaches on varying edge lengths for the proposed uncertainty sets.

\textbf{(i) Constraints on individual edge.} We consider an uncertainty for each edge length $\hat w_e$ of edge $e \in E$ in tree $\calT(\hat w)$. Specifically, we consider $\hat w_e$  belongs to some uncertainty interval $w_e-\alpha_e \le \hat w_e \le w_e +\beta_e$ around the edge weight  $w_e$ in tree $\calT$. In the vector form, this just means that 
\[
w-\alpha \le \hat w \le w + \beta,
\]
with $\alpha, \, \beta\in \RR^{|E|}_{+}$ satisfying $\alpha \leq w$ (i.e., edge weights are nonnegative). Thus, $w-\alpha$ and $w+\beta$ are respectively  the lower and upper limits of the uncertainty interval for the vector of edge lengths.

\textbf{(ii) Constraints on set of edges.} We consider an uncertainty for all edge lengths of tree $\calT(\hat w)$ where the vector of edge lengths $\hat w$ is nonnegative (i.e., $\hat w \ge 0$), and belongs to an uncertainty closed $\ell_p$-ball $\overline\calB_p(w, \lambda)$ centering at $w=(w_e)_{e\in E}$ and with radius $\lambda > 0$. We assume that $1\leq p \leq \infty$ and the uncertainty ball satisfies $\overline\calB_p(w, \lambda)\subset \RR^{|E|}_{+}$.

\paragraph{Closed-form expressions.} By building upon the proposed uncertainty sets of tree metrics and leveraging tree structure, we derive closed-form expressions for the max-min robust OT, similar to its counterpart standard OT (i.e., TW). 

\textbf{$\bullet$ For constraints on individual edge.}
Given two vectors $\alpha, \, \beta\in \RR^{|E|}_{+}$ satisfying $\alpha \leq w$ (to guarantee the nonnegativeness for edge lengths), we define an uncertainty set for tree $\calT(\hat w)$ as follows
\begin{equation*}
\calU(\calT \! , \alpha, \beta) \hspace{-0.2em}  \Let  \hspace{-0.2em}  \left\{ \! \hat \calT \hspace{-0.2em}  = \hspace{-0.2em}  \calT(\hat w)  \hspace{-0.1em} \mid \hspace{-0.1em} - \alpha_e \hspace{-0.1em} \le \hspace{-0.1em} \hat w_e \hspace{-0.1em} - \hspace{-0.1em} w_e \hspace{-0.1em} \le \hspace{-0.1em}  \beta_e, \, \forall e \hspace{-0.1em} \in \hspace{-0.1em} E \right\}.
\end{equation*}
The robust OT in Problem~\eqref{eq:maxmin_robustTW} can be reformulated as
\begin{equation}\label{eq:RTW_local}
\RT_{\calU}(\mu, \nu) 
\Let \max_{\hat\calT \in \calU(\calT, \alpha, \beta)} \calW_{\hat\calT} (\mu, \nu).
\end{equation}
By leveraging the underlying tree structure for OT between measures $\mu$ and $\nu$ with tree metric $d_{\hat \calT}$, we can further rewrite Problem~\eqref{eq:RTW_local} as 
\begin{equation}\label{eq:RTW_edges}
\RT_{\calU}(\mu, \nu) = \max_{w-\alpha \le \hat w \le w + \beta} \sum_{e \in E} \hat w_e \left| \mu(\gamma_e) - \nu(\gamma_e) \right|.
\end{equation}
Notice that $\mu(\gamma_e)$ and $\nu(\gamma_e)$ only depend on the supports of $\mu$ and $\nu$ (i.e., corresponding nodes in $V$) and on the mass on these supports. In particular, these two terms $\mu(\gamma_e)$ and $\nu(\gamma_e)$ are independent of the edge length $\hat w_e$ on each edge $e \in E$ of tree $\calT(\hat w)$. Therefore, we can compute $\RT_{\calU}(\mu, \nu)$ analytically:
\begin{equation}\label{closed-form-local}
\RT_{\calU}(\mu, \nu) = \sum_{e \in E} (w_e + \beta_e) \left| \mu(\gamma_e) - \nu(\gamma_e) \right|,
\end{equation}
where we recall that $w_e + \beta_e$ is the upper limit of the uncertainty edge weight interval for each edge $e \in E$ in $\calU(\calT \! , \alpha, \beta)$.

\textbf{$\bullet$ For constraints on set of edges.} Given a radius $\lambda >0$ such that $\overline\calB_p(w, \lambda)\subset \RR^{|E|}_{+}$, we define an uncertainty set for tree $\calT(\hat w)$ as follows:
\[
\calU_p(\calT, \lambda) \Let \left\{\hat \calT = \calT(\hat w) \mid \hat w \in \overline\calB_p( w, \lambda), \hat w \ge 0 \right\}.
\]
The robust OT corresponding to the uncertainty set $\calU_p(\calT,  \lambda)$ is
\begin{equation}\label{eq:RTW_global}
\RT_{\calU_p}(\mu, \nu) 
\Let \max_{\hat\calT \in \calU_p(\calT, \lambda)} \calW_{\hat\calT} (\mu, \nu).
\end{equation}
Similarly, we leverage the underlying tree structure for OT between $\mu$ and $\nu$ with tree metric $d_{\hat \calT}$ to reformulate the definition in \eqref{eq:RTW_global} as
\begin{equation}\label{eq:RTW_edges_lpball}
\hspace{-0.5em} \RT_{\calU_p}(\mu, \nu) = \max_{\substack{\hat w \ge 0 \\ \hat w \in \overline \calB_p( w, \lambda)}} \sum_{e \in E} \hat w_e \left| \mu(\gamma_e) - \nu(\gamma_e) \right|.
\end{equation}
By simply leveraging the dual norm, we derive the closed-form expression for $\RT_{\calU_p}(\mu, \nu)$ in Problem \eqref{eq:RTW_edges_lpball}:
\begin{proposition}\label{prop:globalRTW_closedform}
Assume that $1\leq p \leq \infty$. Then, 
\begin{equation}\label{eq:RTW_lpball}
\hspace{-0.6em} \RT_{\calU_p}(\mu, \nu) = \Big(\sum_{e \in E}  w_e \left| \mu(\gamma_e) - \nu(\gamma_e) \right|\Big) + \lambda \norm{h}_{p'}, 
\end{equation}
where $h \in \RR^{|E|}$ is the vector with $h_e \Let\left| \mu(\gamma_e) - \nu(\gamma_e) \right|$ for each edge $e$, and $p'$ is the conjugate of $p$. Moreover, a maximizer for Problem \eqref{eq:RTW_edges_lpball} is given by $\hat w_e^* =  w_e + \lambda \|h\|_{p'}^{-\frac{p'}{p}} h_e^{p' -1}$ for $1 < p < \infty$; $\hat w_e^* =  w_e + \lambda$ for $p=\infty$; and for $p=1$, let $e^*\in E$ be s.t. $\|h\|_{\infty} = h_{e^*} > 0$, then
\[
\hat w_e^*  =
\left\{\begin{array}{lr}
\!\!w_{e^*} + \lambda   \hspace{1 em} \mbox{ if } \, e = e^*,\\
\!\!w_e \hspace{3.3 em} \mbox{ otherwise.}
\end{array}\right.
\]
\end{proposition}
To our knowledge, among various approaches for the max-min robust OT in the literature, our proposed approach is the \emph{first} one which yields a closed-form expression for fast computation, and is scalable for large-scale applications. 

\paragraph{Connection between two approaches.} We next draw a connection between two approaches for the robust OT for measures with noisy tree metric as follows:
\begin{proposition}[Connection between two approaches]\label{prop:connection_global_local}
Assume that $\beta = \lambda \mathds{1}$. Then, we have
\begin{equation}\label{eq:connection_global_local}
\RT_{\calU(\calT, \alpha, \beta)}(\cdot, \cdot) = \RT_{\calU_\infty(\calT, \lambda)}(\cdot, \cdot).
\end{equation}
\end{proposition}


\paragraph{Computational complexity.} From the closed-form expressions for $\RT_{\calU}$ in~\eqref{closed-form-local} and for $\RT_{\calU_p}$ in~\eqref{eq:RTW_lpball}, the computational complexity of robust OT $\RT_{\calU}$ and $\RT_{\calU_p}$ for measures with noisy tree metric is linear to the number of edges in $\calT$ (i.e., $\mathcal{O}(|E|)$), which is in the same order of computational complexity as the counterpart standard OT for measures with tree metric (i.e., TW)~\citep{ba2011sublinear, LYFC}. Recall that, in general, the max-min robust OT problem is hard and expensive to compute due to its non-convexity and non-smoothness~\citep{pmlr-v97-paty19a, lin2020projection}, even for measures supported in one-dimensional space~\citep{deshpande2019max}.\footnote{Even with a given optimal ground metric cost, the computational complexity of max-min/min-max robust OT is in the same order as their counterpart standard OT (i.e., their objective function).}

\textbf{Improved complexity.} Let $\text{supp}(\mu)$ and $\text{supp}(\nu)$ be supports of measures $\mu$ and $\nu$ respectively, and define 
\[
E_{\mu, \nu} \hspace{-0.15em} \Let \hspace{-0.15em} \left\{e \in E \mid e \subset [r, z] \mbox{ with $z \in$ }\text{supp}(\mu) \cup \text{supp}(\nu) \right\}.
\]
Then, observe that $\mu(\gamma_e) = \nu(\gamma_e) = 0$ for any edge $e \in E \setminus E_{\mu, \nu}$. Consequently, we can further reduce the computational complexity of the robust OT $\RT_{\calU}$ and $\RT_{\calU_p}$ into just $\mathcal{O}(|E_{\mu, \nu}|)$. 



\paragraph{Negative definiteness.} We next prove that the robust OT for measures with noisy tree metric is negative definite. Therefore, we can derive positive definite kernels built upon the robust OT. 
\begin{theorem}[Negative definiteness]\label{thrm:negative_definite}
$\RT_{\calU}$ is negative definite.
In addition,  $\RT_{\calU_p}$ is also negative definite for all $ 2\leq p \leq \infty$.
\end{theorem}

\textbf{Positive definite kernels.} From the negative definite results in Theorem~\ref{thrm:negative_definite} and by following \citep[Theorem~3.2.2, pp.74]{Berg84}, given $2\le p\le \infty$ and $t > 0$, we propose positive definite kernels built upon the robust OT for both approaches as follows:
\begin{eqnarray}\label{eq:kernel_RTW}
& k_{\RT_{\calU}}(\mu, \nu) = \exp(-t \RT_{\calU}(\mu, \nu)), \label{eq:kernel_RTW_Local} \\
& k_{\RT_{\calU_p}}(\mu, \nu) = \exp(-t \RT_{\calU_p}(\mu, \nu)) \label{eq:kernel_RTW_Global}.
\end{eqnarray}
To our knowledge, among various existing approaches for the max-min/min-max robust OT, our work is the \emph{first} provable approach to derive positive definite kernels built upon the robust OT.\footnote{In general, Wasserstein space is \emph{not} Hilbertian~\citep[\S8.3]{peyre2019computational}, and the standard OT is indefinite. Thus, it is nontrivial to build positive definite kernels upon OT for probability measures.}

\textbf{Infinite divisibility for the robust OT kernels.} We next illustrate the infinite divisibility for the robust OT kernels for measures with noisy tree metric.
\begin{proposition}[Infinitely divisible kernels]\label{prop:divisibility}
The kernel $k_{\RT_{\calU}}$ is infinitely divisible. Also, the kernel $k_{\RT_{\calU_p}}$ is infinitely divisible for all $2\le p\le \infty$.
\end{proposition}
As for infinitely divisible kernels, one does not need to recompute the Gram matrix of kernels $k_{\RT_{\calU}}$ and $k_{\RT_{\calU_p}}$ with $2\le p\le \infty$ for each choice of hyperparameter $t$, since it suffices to compute these robust OT kernels for probability measures in the training set once. 

\paragraph{Metric property.} We end this section by showing that the robust OT is a metric.
\begin{proposition}[Metric]\label{prop:metric}
    $\RT_{\calU}$ is a metric. Also, $\RT_{\calU_p}$ is  a metric for all $1 \le p \le \infty$.
\end{proposition}

\section{RELATED WORK AND DISCUSSION}\label{sec:related_work}

In this section, we discuss related work to the max-min robust OT approach for OT problem for measures with noisy tree metric. We further distinguish it with other lines of research in OT.

One of seminal works in max-min robust OT is the projection robust Wasserstein~\citep{pmlr-v97-paty19a} (i.e., Wasserstein projection pursuit~\citep{niles2022estimation}). This approach considers the maximal possible Wasserstein distance over all possible low dimensional projections. This problem is non-convex and non-smooth, which is hard and expensive to compute~\citep{lin2020projection}. By leveraging the Riemannian optimization, \citet{lin2020projection} derived an efficient algorithmic approach which provides the finite-time guarantee for the computation of the projection robust Wasserstein. \citet{pmlr-v97-paty19a} considered its convexified relaxation min-max robust OT, namely subspace robust Wasserstein distance, which provides an upper bound for the projection robust Wasserstein.~\citet{alvarez2018structured} proposed the submodular OT to reflect additional structures for OT.~\citet{pmlr-v84-genevay18a} used min-max robust OT as a loss to improve generative model for images.~\citet{dhouib2020swiss} considered the minimax OT which jointly optimizes the cost matrix and the transportation plan for OT.


Notice that the robust OT approach for OT problem with noisy ground cost is dissimilar to the Wasserstein distributionally robust optimization~\citep{kuhn2019wasserstein, blanchet2021statistical}. Although they may share the min-max formulation, the Wasserstein distributionally robust optimization seeks the best data-driven decision under the most adverse distribution from a Wasserstein ball of a certain radius. Additionally, one should distinguish this robust OT approach for noisy ground cost with the outlier-robust approach for OT where the noise is on input probability measures~\citep{balaji2020robust, mukherjee2021outlier, nguyen2021robust, nietert2022outlier}.

 
Leveraging tree structure to scale up OT problems has been explored for standard OT~\citep{LYFC, yamada2022approximating}, for OT problem with input measures having different total mass~\citep{sato2020fast, le2021ept}, and for Wasserstein barycenter~\citep{pmlr-v151-takezawa22a}. To our knowledge, our work is the \emph{first} approach to exploit tree structure over supports to scale up robust OT approach for OT problem with noisy ground cost. Furthermore, notice that max-sliced Wasserstein~\citep{deshpande2019max} is one-dimensional OT-based approach for the max-min robust OT. However, there are no fast/efficient algorithmic approaches yet due to its non-convexity. Our approach is based on tree structure which provides more flexibility and degrees of freedoms to capture the topological structure of input probability measures than the one-dimensional OT-based approach (i.e., choosing a tree rather than a line). Moreover, our novel uncertainty sets of tree metrics play the key role to scale up the computation of robust OT. The uncertainty sets not only includes a diversity of tree structures, but also a variety of edge lengths in an elegant framework following the theoretical guidance in Theorem~\ref{thrm:zeroedges}.


\begin{figure*}
 \vspace{-4pt}
  \begin{center}
    \includegraphics[width=0.9\textwidth]{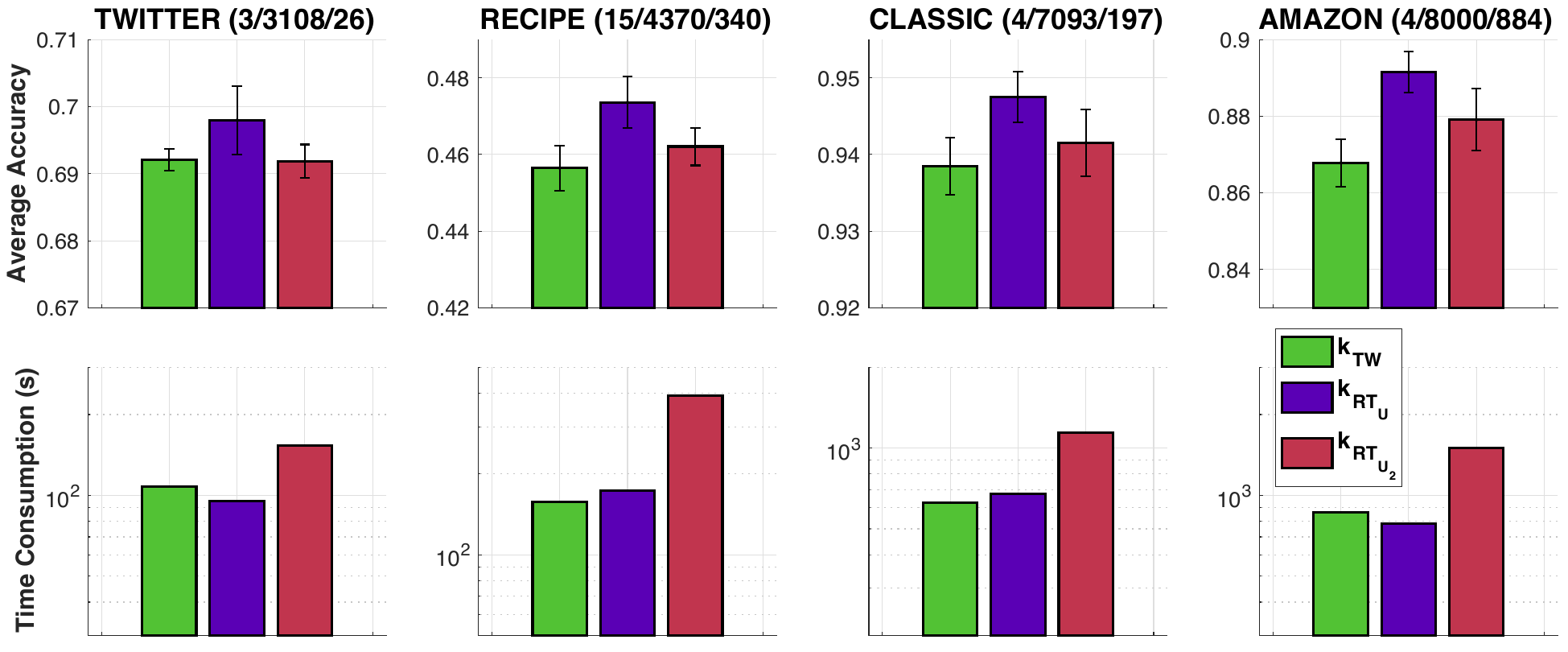}
  \end{center}
  \vspace{-12pt}
  \caption{SVM results and time consumption for kernel matrices in document classification. For each dataset, the numbers in the parenthesis are the number of classes; the number of documents; and the maximum number of unique words for each document respectively.}
  \label{fg:DOC_Noise_Delta0.5}
 \vspace{-4pt}
\end{figure*}

\section{EXPERIMENTS}\label{sec:experiments}


In this section, we illustrate: (i) fast computation for $\RT_{\calU}$ and $\RT_{\calU_p}$, (ii) the robust OT kernels $k_{\RT_{\calU}}$ and $k_{\RT_{\calU_p}}$ improve performances of the counterpart standard OT (i.e., tree-Wasserstein) kernel \emph{for measures} with noisy tree metric, similar to other existing max-min/min-max robust OT in the OT literature.

More concretely, we compare the proposed robust OT kernels $k_{\RT_{\calU}}$ and $k_{\RT_{\calU_p}}$ with the counterpart standard OT (i.e., TW) kernel $k_{\TW}$, defined as $k_{\TW}(\cdot, \cdot) = \exp(-t \calW_{\calT}(\cdot, \cdot))$ for a given $t > 0$, for measures with a given noisy tree metric under the same settings on several simulations on document classification and topological data analysis (TDA) with SVM.\footnote{One may not directly use existing robust OT approaches with Euclidean geometry for measures with a given noisy tree metric since the considered problem does not satisfy such conditions.} 

We emphasize there are various approaches for the simulations on document classification and TDA. However, it is not the goal of our study.

\paragraph{Document classification.} We evaluate on $4$ real-world document datasets: \texttt{TWITTER}, \texttt{RECIPE}, \texttt{CLASSIC}, and \texttt{AMAZON}. Their characteristics are listed in Figure~\ref{fg:DOC_Noise_Delta0.5}. We follow the same approach in~\citep{LYFC} to embed words into vectors in $\R^{300}$, and represent each document as a probability measure where its supports are in $\R^{300}$.



\begin{figure}
  \begin{center}
    \includegraphics[width=0.45\textwidth]{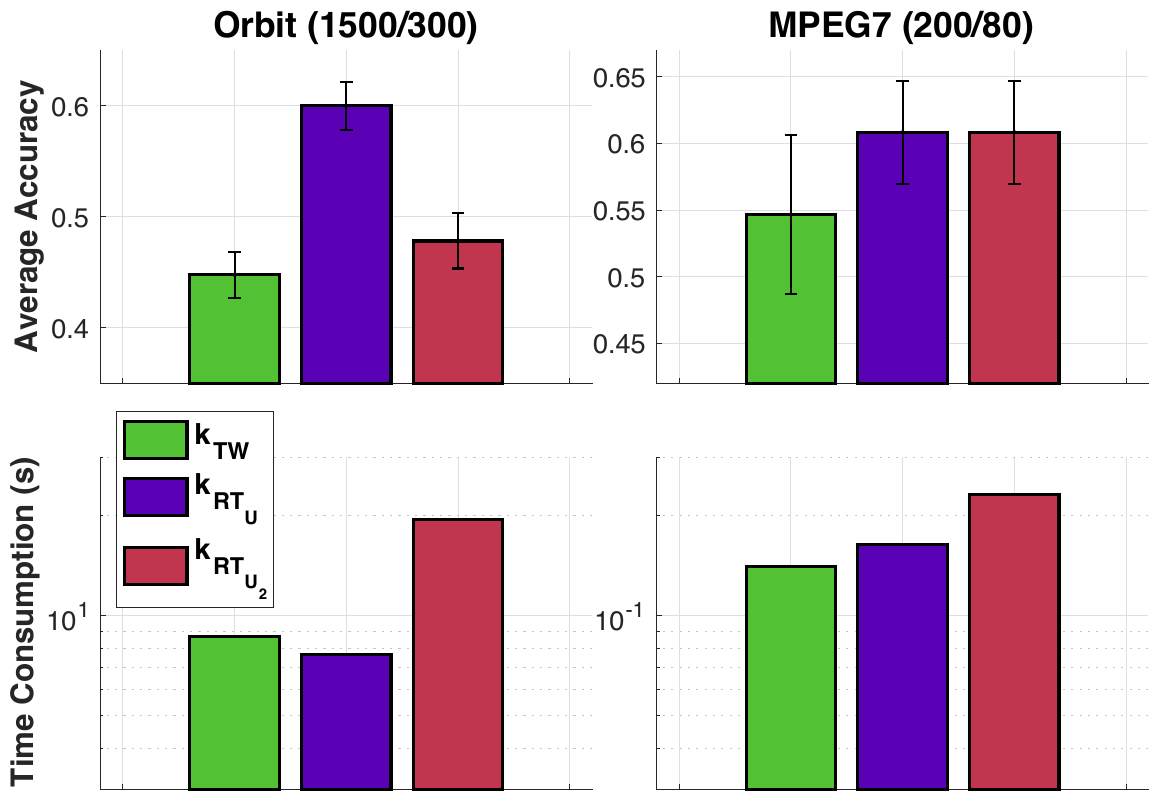}
  \end{center}
  \vspace{-12pt}
  \caption{SVM results and time consumption for kernel matrices in TDA. For each dataset, the numbers in the parenthesis are respectively the number of PD; and the maximum number of points in PD.}
  \label{fg:TDA_Noise_Delta0.05}
 \vspace{-4pt}
\end{figure}

\paragraph{TDA.} We consider orbit recognition on \texttt{Orbit} dataset and object shape recognition on \texttt{MPEG7} dataset for TDA. We summarize the characteristics of these datasets in Figure~\ref{fg:TDA_Noise_Delta0.05}. We follow the same approach in~\citep{LYFC} to extract persistence diagram (PD) for orbits and object shapes, which are multisets of points in $\R^2$, and represent each PD as a probability measure where its supports are in $\R^2$.

\paragraph{Noisy tree metric.} We apply the clustering-based tree metric sampling method~\citep{LYFC} to obtain a tree metric over supports. We then generate perturbations by deviating each tree edge length by a random nonnegative amount which is less than or equal $\Delta \in \RR_+$ (i.e., $|w_e - w_e^*| \le \Delta$) where $w_e^*, w_e$ are tree edge lengths on the tree before and after the perturbations respectively. We set $\Delta=0.5$ for document classification (for edge lengths constructed from supports in $\RR^{300}$); and set $\Delta=0.05$ for TDA (for edge lengths constructed from supports in $\RR^{2}$). 

Following Theorem~\ref{thrm:zeroedges}, the perturbations suffice to cover various tree structures via $0$-length edges (i.e., all subtree structures of the tree before perturbations). Moreover, it is not necessary to add more $0$-length edges before the perturbations for our experiments.\footnote{E.g., in Figure~\ref{fg:tree_transform}, if we add edge $e_b$ (as in $\calT_3$ from $\calT_1$), there are no supports on node $v_b$ for any input measures; and if we add edge $e_a$ (as in $\calT_3$ from $\calT_1$), it is equivalent to perturb edge $e_1$ in $\calT_1$ by the total amount of perturbations on edges $e_1$ and $e_a$ in $\calT_3$.}





\begin{figure*}
  \vspace{-4pt}
  \begin{center}
    \includegraphics[width=0.9\textwidth]{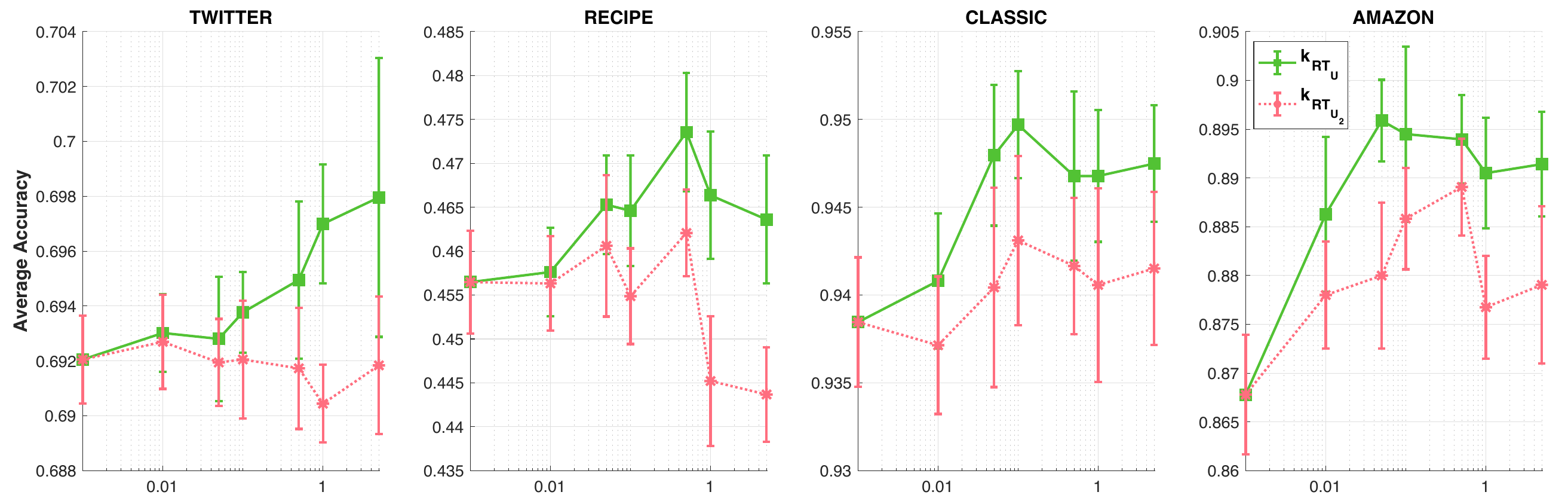}
  \end{center}
  \vspace{-12pt}
  \caption{SVM results for document classification w.r.t. the radius $\lambda$.} 
  \label{fg:DOC_ParaAlpha_Noise_Delta0.5}
 \vspace{-4pt}
\end{figure*}

\paragraph{Experimental setup.} We apply kernel SVM for the proposed robust OT kernels $k_{\RT_{\calU}}$ and $k_{\RT_{\calU_{p}}}$ and the counterpart standard OT (i.e., TW) kernel $k_{\TW}$ for measures with a given noisy tree metric on document classification and TDA. For $\RT_{\calU}$, we consider $\alpha = \min(\lambda \mathds{1}, w)$ and $\beta = \lambda \mathds{1}$ where minimum operator is element-wise; $\lambda$ is the radius in $\RT_{\calU_{p}}$; and recall that $w$ is a vector of edge lengths of the given tree. We set $p=2$ for the robust OT $\RT_{\calU_{p}}$ (or $\RT_{\calU_{2}}$).

For kernel SVM, we use one-versus-one approach for SVM with multiclass data points. We randomly split each dataset into $70\%/30\%$ for training and test with $10$ repeats. Typically, we use cross validation to choose hyperparameters. For the kernel hyperparameter $t$, we choose $1/t$ from $\{q_{s}, 2q_{s}, 5q_{s}\}$ with $s \!=\! 10, 20, \dotsc, 90$, where $q_s$ denotes the $s\%$ quantile of a random subset of corresponding distances on training data. For SVM regularization hyperparameter, we choose it from $\left\{0.01, 0.1, 1, 10, 100\right\}$. For the radius $\lambda$ in $\RT_{\calU_{2}}$ (also in $\RT_{\calU}$ through the choice of $\beta$), we choose it from $\left\{0.01, 0.05, 0.1, 0.5, 1, 5\right\}$. All our experiments are run on commodity hardware.


 \begin{figure}
  \begin{center}
    \includegraphics[width=0.45\textwidth]{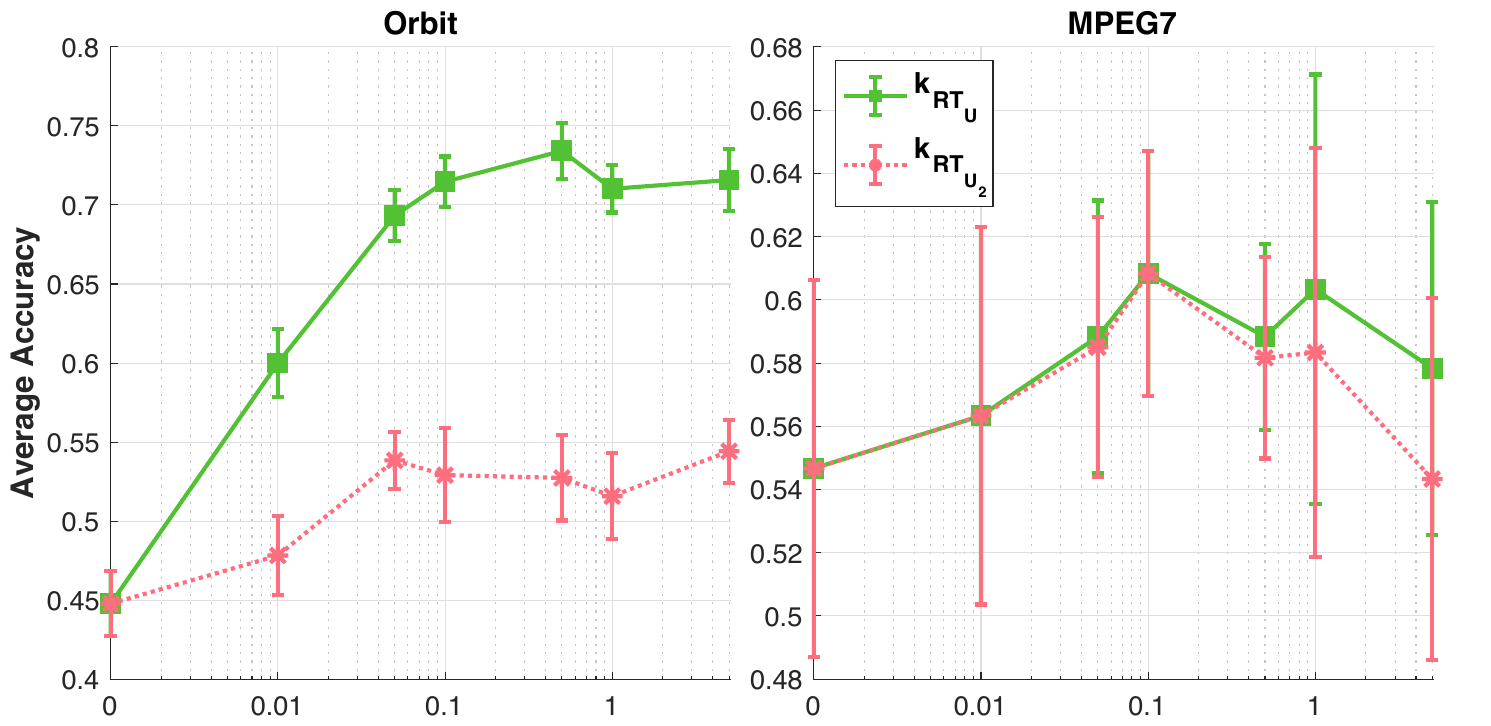}
  \end{center}
  \vspace{-12pt}
  \caption{SVM results for TDA w.r.t. the radius $\lambda$.}
  \label{fg:TDA_ParaAlpha_Noise_Delta0.05}
 \vspace{-4pt}
\end{figure}

\paragraph{Empirical results and discussion.} Figures~\ref{fg:DOC_Noise_Delta0.5} and~\ref{fg:TDA_Noise_Delta0.05} illustrate the performances of SVM for document classification and TDA respectively. The performances of the proposed kernels $k_{\RT_{\calU}}$ and $k_{\RT_{\calU_2}}$ compare favorably to those of the counterpart OT kernel $k_{\TW}$. Notably, kernel $k_{\RT_{\calU}}$ improves about 15\% average accuracy over kernel $k_{\TW}$ on \texttt{Orbit}. In addition, kernel $k_{\RT_{\calU}}$ consistently outperforms kernel $k_{\RT_{\calU_2}}$, except on \texttt{MPEG7} where their performances are comparable, which may come from the freedom to constraint over each edge length in $\calU$ of kernel $k_{\RT_{\calU}}$. Moreover, our results also agree with previous observations on other existing max-min/min-max robust OT where the robust OT approach improves performances of the counterpart standard OT for measures with noisy ground cost. For further discussions, we refer the readers to the supplementary (\S C--\S D).


Furthermore, the kernels $k_{\RT_{\calU}}$ and $k_{\RT_{\calU_2}}$ are fast for computation. They are in the same order as that of the counterpart standard OT (i.e., TW) kernel for measures with noisy tree metric, which agrees with our theoretical analysis in \S\ref{sec:RobustTW} (i.e., linear to the number of edges in the given tree).\footnote{The computational complexity of OT is in general super cubic w.r.t. the number of supports of input measures. We also refer the readers to~\citep{LYFC} for extensive results about the trade-off between performances and time consumptions for the standard OT, TW and SW.} This is in stark contrast with other existing max-min (or even min-max) robust OT since it is already costly to evaluate the objective function, which is a standard OT for a fixed ground cost, besides the hardness of non-convex and non-smooth optimization problem for max-min robust OT in general.

We next investigate effects of radius $\lambda$ for the robust OT. Recall that $\lambda$ is the radius of the $\ell_2$-ball uncertainty for $\RT_{\calU_2}$, and we use $\lambda$ for parameter $\beta$ in $\RT_{\calU}$.

\paragraph{Effects of the radius \boldsymbol{$\lambda$}.} Figures~\ref{fg:DOC_ParaAlpha_Noise_Delta0.5} and~\ref{fg:TDA_ParaAlpha_Noise_Delta0.05} illustrate the effects of the radius $\lambda$ on the proposed robust OT kernels on document classification and TDA respectively. Notice that when $\lambda = 0$, the max-min robust OT for are equivalent to the counterpart standard OT. We observe that kernel $k_{\RT_{\calU}}$ is less sensitive with the radius $\lambda$ than kernel $k_{\RT_{\calU_2}}$. The performances of kernel $k_{\RT_{\calU}}$ gradually increase when the radius $\lambda$ increases, after these performances reach their peaks, they decrease when $\lambda$ increases. The performances of kernel $k_{\RT_{\calU_2}}$ also share a similar pattern but more noisy. Therefore, cross validation for the radius $\lambda$ is useful in applications, especially for kernel $k_{\RT_{\calU_2}}$ in our simulations.

We place further empirical results with different parameters in the supplementary (\S D).


\section{CONCLUSION}\label{sec:conclusion}

In this work, we proposed novel uncertainty sets of tree metrics which not only include metric metrics with varying edge lengths, but also having diverse tree structures in an elegant framework. By building upon these uncertainty sets and leveraging tree structure, we scale up the max-min robust OT approach for OT problem for probability measures with noisy tree metric. Moreover, by exploiting the negative definiteness of the robust OT, we proposed positive definite kernels built upon the robust OT and evaluated them for kernel SVM on document classification and TDA. For future work, extending the problem settings for more general applications (e.g., by leveraging the clustering-based tree metric sampling method~\citep{LYFC}), and/or for more general structures (e.g., graphs as in the Sobolev transport~\citep{le2022sobolev, le2023scalable}) are interesting research directions.


\subsubsection*{Acknowledgements}
We thank the area chairs and anonymous reviewers for their comments. KF has been supported in part by Grant-in-Aid for Transformative Research Areas (A) 22H05106. TL gratefully acknowledges the support of JSPS KAKENHI Grant number 23K11243.

\balance

\bibliography{bibEPT21,bibSobolev22}
\bibliographystyle{icml2022}

\section*{Checklist}

 \begin{enumerate}

 \item For all models and algorithms presented, check if you include:
 \begin{enumerate}
   \item A clear description of the mathematical setting, assumptions, algorithm, and/or model. [Yes]
   \item An analysis of the properties and complexity (time, space, sample size) of any algorithm. [Yes]
   \item (Optional) Anonymized source code, with specification of all dependencies, including external libraries. [Yes]
 \end{enumerate}

 \item For any theoretical claim, check if you include:
 \begin{enumerate}
   \item Statements of the full set of assumptions of all theoretical results. [Yes]
   \item Complete proofs of all theoretical results. [Yes]
   \item Clear explanations of any assumptions. [Yes]     
 \end{enumerate}

 \item For all figures and tables that present empirical results, check if you include:
 \begin{enumerate}
   \item The code, data, and instructions needed to reproduce the main experimental results (either in the supplemental material or as a URL). [Yes]
   \item All the training details (e.g., data splits, hyperparameters, how they were chosen). [Yes]
         \item A clear definition of the specific measure or statistics and error bars (e.g., with respect to the random seed after running experiments multiple times). [Yes]
         \item A description of the computing infrastructure used. (e.g., type of GPUs, internal cluster, or cloud provider). [Yes]
 \end{enumerate}

 \item If you are using existing assets (e.g., code, data, models) or curating/releasing new assets, check if you include:
 \begin{enumerate}
   \item Citations of the creator If your work uses existing assets. [Yes]
   \item The license information of the assets, if applicable. [Not Applicable]
   \item New assets either in the supplemental material or as a URL, if applicable. [Yes]
   \item Information about consent from data providers/curators. [Yes]
   \item Discussion of sensible content if applicable, e.g., personally identifiable information or offensive content. [Not Applicable]
 \end{enumerate}

 \item If you used crowdsourcing or conducted research with human subjects, check if you include:
 \begin{enumerate}
   \item The full text of instructions given to participants and screenshots. [Not Applicable]
   \item Descriptions of potential participant risks, with links to Institutional Review Board (IRB) approvals if applicable. [Not Applicable]
   \item The estimated hourly wage paid to participants and the total amount spent on participant compensation. [Not Applicable]
 \end{enumerate}

 \end{enumerate}

\appendix
\onecolumn



\begin{center}
{\bf{\Large{Supplement to  ``Optimal Transport for Measures with Noisy Tree Metric"}}}
\end{center}

In this supplementary, we give brief reviews about some aspects used in our work, e.g.,  kernels, tree metric, and optimal transport on for probability measures on a tree in \S\ref{appsec:review}. We  present detailed proofs for the theoretical results in \S\ref{appsec:proofs}, and give additional discussions about our work in \S\ref{appsec:discussion}. Further experimental results are placed in \S\ref{appsec:experiment}.

\section{BRIEF REVIEWS}
\label{appsec:review}

In this section, we briefly review about some aspects used in our work.

\subsection{Kernels} 

We review definitions and theorems about kernels that are used in our work.

\paragraph{Positive Definite Kernels \cite[pp.~66--67]{Berg84}.} A kernel function $k: \Omega \times \Omega \rightarrow \R$ is positive definite if for every positive integer $m \ge 2$ and every points $x_1, x_2, ..., x_m \in \Omega$, we have 
\[
\sum_{i, j=1}^m c_i c_j k(x_i, x_j) \ge 0 \qquad \forall c_1,...,c_m \in \R.
\]

\paragraph{Negative Definite Kernels \cite[pp.~66--67]{Berg84}.} A kernel function $k: \Omega \times \Omega \rightarrow \R$ is negative definite if for every integer  $ m \ge 2$ and every points $x_1, x_2, ..., x_m \in \Omega$, we have 
\[
\sum_{i, j=1}^m c_i c_j k(x_i, x_j) \le 0, \qquad \quad \forall c_1,...,c_m \in \R\,\, \text{ s.t. } \, \sum_{i=1}^m c_i = 0.
\]

\paragraph{Theorem 3.2.2 in \citet[pp.~74]{Berg84}.}
Let  $\kappa$ be a \textit{negative definite} kernel function. Then, for every $t>0$, the  kernel 
\[
k(x, z) \Let \exp{\left(- t \kappa(x, z)\right)}
\]
is positive definite.

\paragraph{Definition 2.6 in \citet[pp.~76]{Berg84}.}
A positive definite kernel $\kappa$ is \emph{infinitely divisible} if for each $n \in {\N}^{*}$, there exists a positive definite kernel $\kappa_n$ such that 
\[
\kappa = (\kappa_n)^n.
\]

\paragraph{Corollary 2.10 in \citet[pp.~78]{Berg84}.}
Let  $\kappa$ be a \textit{negative definite} kernel function. Then, for $0 < t < 1$, the  kernel 
\[
k(x, z) \Let \left[\kappa(x, z)\right]^{t}
\]
is negative definite.

\subsection{Tree Metric} 

We review the definition of tree metric and give detailed references for the clustering-based tree metric sampling method used in our experiments.

\paragraph{Tree metric.} A metric $d:\Omega\times\Omega\rightarrow \mathbf{R}$ is a tree metric on $\Omega$ if there exists a tree $\calT$ with non-negative edge lengths such that all elements of $\Omega$ are contained in its nodes and such that for every $x, z \in \Omega$, we have $d(x, z)$ equals to the length of the path between $x$ and $z$~\citep[\S7, pp.145--182]{semple2003phylogenetics}. We write $d_{\calT}$ for the tree metric corresponding to the tree $\calT$.

\paragraph{Clustering-based tree metric sampling method.} The clustering-based tree metric sampling method was proposed by \citet{LYFC} (see their \S4). Additionally, \citet{LYFC} reviewed the farthest-point clustering in \S4.2 in their supplementary, which is the main component used in the clustering-based tree metric sampling method.

\subsection{Optimal Transport (OT) for Measures on a Tree}

\begin{figure}[h]
  \begin{center}
    \includegraphics[width=0.35\textwidth]{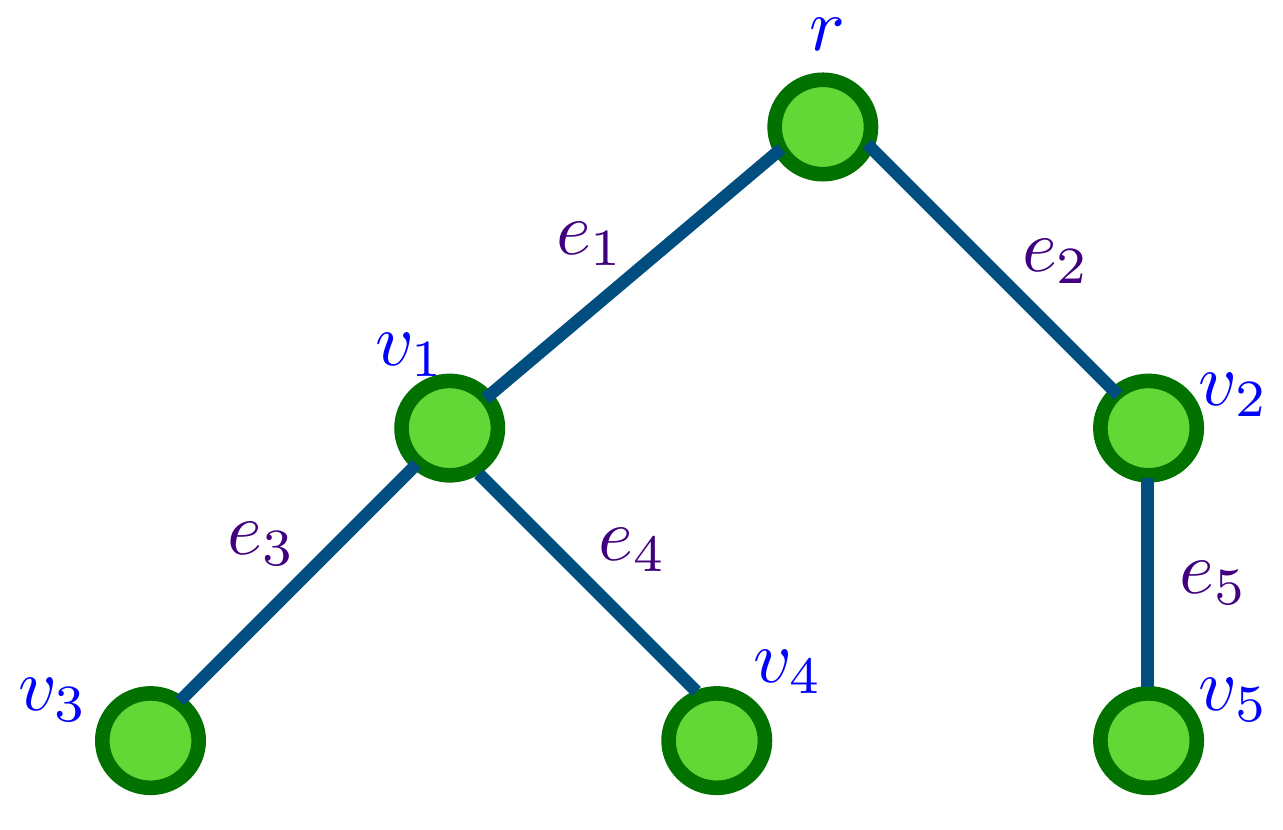}
  \end{center}
  \caption{An illustration for a tree with root $r$. The set of nodes $V = \{ r, v_1, v_2, v_3, v_4, v_5\}$ and the set of edges $E = \{e_1, e_2, e_3, e_4, e_5 \}$. For the edge $e_1$, we have $\gamma_{e_1} = \{v_1, v_3, v_4 \}$. The path $[r, v_3] = \{e_1, e_3 \}$.}
  \label{fg:tree}
\end{figure}

The $1$-Wasserstein distance $\calW_{\calT}$ (with tree metric $d_{\calT}$ as its ground cost), i.e., tree-Wasserstein (TW), admits a closed-form expression~\citep{ba2011sublinear, LYFC}
\begin{equation}\label{eq:TW}
\calW_{\calT}(\mu, \nu) =\sum_{e \in E} w_e \left|\mu(\gamma_e) - \nu(\gamma_e) \right|,
\end{equation}
where $\mu, \, \nu \in \calP(V)$ and recall that $\gamma_e$ is the set of all nodes $x$ such that the path $ [r, x]$ contains the edge $e$. An illustration of a set $\gamma_e$ is given in Figure~\ref{fg:tree}.

Thanks to formula \eqref{eq:TW}, the computational complexity of TW is linear to the number of edges on tree $\calT$. Recall that in general the computational complexity of OT is super cubic w.r.t. the number of supports of input measures. We refer the readers to~\citep{LYFC} for further details of TW.

\subsection{Persistence Diagrams and Definitions in Topological Data Analysis.} 

We refer the readers to \citet[\S2]{kusano2017kernel} for a brief review about the mathematical framework for persistence diagrams (e.g., persistence diagrams, filtrations, persistent homology).

\section{PROOFS}\label{appsec:proofs}

In this section, we give proofs for the theoretical results in the main manuscript.

\subsection{Proof for Theorem~\ref{thrm:zeroedges}}

\begin{proof}


Let $E$ and $E'$ be the set of edges on trees $\calT$ and  $\calT'$ respectively. Let $e$ be the $0$-length edge which we collapse from tree $\calT$ to construct tree $\calT'$. Then, we have 
\[
E = E' \cup \{e\}.
\]
Moreover, observe that for any edge $\tilde{e} \in E'$, $\mu_{\calT'}(\gamma_{\tilde{e}})$ and $\nu_{\calT'}(\gamma_{\tilde{e}})$ in the constructed tree $\calT'$ are the same as their corresponding ones $\mu_{\calT}(\gamma_{\tilde{e}})$ and $\nu_{\calT}(\gamma_{\tilde{e}})$ in the original tree $\calT$ respectively. Therefore, from formula \eqref{eq:TW} and since $w_e = 0$, we obtain 
\begin{align*}
\calW_{\calT}(\mu, \nu) 
&=\sum_{\tilde e \in E} w_{\tilde e} \left|\mu(\gamma_{\tilde e}) - \nu(\gamma_{\tilde e}) \right|\\
&=\Big(\sum_{{\tilde e} \in E'} w_{\tilde e} \left|\mu_{\calT'}(\gamma_{\tilde e}) - \nu_{\calT'}(\gamma_{\tilde e}) \right|\Big) +  w_{ e} \left|\mu(\gamma_{ e}) - \nu(\gamma_{ e}) \right|\\
&=\sum_{e \in E'} w_e \left|\mu_{\calT'}(\gamma_e) - \nu_{\calT'}(\gamma_e) \right| = \calW_{\calT'}(\mu_{\calT'}, \nu_{\calT'}).
\end{align*}
This completes the proof.



\end{proof}

\subsection{Proof for Proposition~\ref{prop:globalRTW_closedform}} 

\begin{proof}
Recall that  $h \in \RR^{|E|}$ is given by $h_e = \left| \mu(\gamma_e) - \nu(\gamma_e) \right|$, and as pointed out right after \eqref{eq:RTW_edges} that it is independent of  the edge length $\hat w_e$ on each edge $e \in E$ in tree $\calT(\hat w)$. The conclusion of identity \eqref{eq:RTW_lpball} is obvious if $h$ is the zero vector, and hence we only need to consider the case $h\neq 0$. 

Let consider the following problem:
\begin{equation}\label{eq:RTW_edges_lpball_woNon}
\widetilde \RT_{\calU_p}(\mu, \nu) = \max_{\hat w \in \overline \calB_p( w, \lambda)} \sum_{e \in E} \hat w_e \left| \mu(\gamma_e) - \nu(\gamma_e) \right|,
\end{equation}
which is similar as $\RT_{\calU_p}(\mu, \nu)$ in Problem~\eqref{eq:RTW_edges_lpball}, but without the nonnegative constraint on $\hat w$ (i.e., $\hat w \ge 0$). We will show that the optimal solution $\hat w^*$ in Problem~\eqref{eq:RTW_edges_lpball_woNon} is nonnegative. Hence,  $\widetilde \RT_{\calU_p}(\mu, \nu) = \RT_{\calU_p}(\mu, \nu)$.

Due to the continuity, we have
\begin{equation*}
\widetilde \RT_{\calU_p}(\mu, \nu) = \max_{\hat w \in \overline\calB_p( w, \lambda)} \sum_{e \in E} \hat w_e h_e.
\end{equation*}
This can be further expressed as 
\begin{equation}\label{translate}
\widetilde \RT_{\calU_p}(\mu, \nu) 
 =  \sum_{e \in E}   w_e h_e 
+ 
 \max_{\hat w \in \overline\calB_p(w, \lambda)}  \sum_{e \in E}  (\hat w_e -  w_e)  h_e.
\end{equation}
Since 
$
\sum_{e \in E} (\hat w_e -  w_e)  h_e
\leq \sum_{e \in E} |\hat w_e -  w_e|  h_e
\leq \|\hat w -  w\|_p \|h\|_{p'},
$
we have on one hand that
\begin{align}\label{upperbound-p'}
 \max_{\hat w \in \overline\calB_p( w, \lambda)} \sum_{e \in E} (\hat w_e - w_e)  h_e 
 \leq \lambda \|h\|_{p'}.
\end{align}


On the other hand, for the case $1 < p <\infty$, by taking 
\[
\hat w_e^* \Let  w_e + \lambda \|h\|_{p'}^{-\frac{p'}{p}} h_e^{p' -1}
\]
and as $p'=\frac{p}{p-1}$  we see that $\|  \hat w^* -  w\|_p = \lambda$ and
\begin{align*}
  \sum_{e \in E} (\hat w_e^* -  w_e)  h_e 
 &=\lambda \|h\|_{p'}^{-\frac{p'}{p}} \sum_{e \in E}  h_e^{p'}\\
 &= \lambda \|h\|_{p'}^{p'(1-\frac1p)} = \lambda \|h\|_{p'}.
\end{align*}
Therefore, we conclude that 
\begin{align*}
 \max_{\hat w \in \overline\calB_p( w, \lambda)} \sum_{e \in E} (\hat w_e -  w_e)  h_e =\lambda \|h\|_{p'}
\end{align*}
with $\hat w = \hat w^*$ being a maximizer. Additionally, notice that $w, h \ge 0$, hence, $\hat w^* \ge 0$. Therefore, together with \eqref{translate} yields the conclusion of the Proposition for the case $1 < p <\infty$.

For the case $p=1$, let $e^*\in E$ be such that 
$\|h\|_{\infty} = h_{e^*} > 0$. Then, by taking 
\begin{equation*}
\hat w_e^*  \Let
\left\{\begin{array}{lr}
\!\!w_{e^*} + \lambda   \hspace{1 em} \mbox{ if } \, e = e^*,\\
\!\!w_e \hspace{3.3 em} \mbox{ otherwise,}
\end{array}\right.
\end{equation*}
we see that $\|  \hat w^* -  w\|_1 = \lambda$ and
\begin{align*}
  \sum_{e \in E} (\hat w_e^* -  w_e)  h_e 
 &=\lambda h_{e^*}  = \lambda \|h\|_{\infty}.
\end{align*}
This and \eqref{upperbound-p'} imply that $\max_{\hat w \in \overline\calB_p( w, \lambda)} \sum_{e \in E} (\hat w_e -  w_e)  h_e =\lambda \|h\|_{\infty}$ with $\hat w = \hat w^*$ being a maximizer, and notice that $\hat w^* \ge 0$. The conclusion of the Proposition for the case $p=1$ then follows from this and \eqref{translate}.

For the case $p=\infty$, 
by taking 
$\hat w_e^* 
\Let  w_e + \lambda$ for $e\in E$, 
we see that $\|  \hat w^* -  w\|_\infty = \lambda$ and
\begin{align*}
  \sum_{e \in E} (\hat w_e^* -  w_e)  h_e 
 &= \lambda \sum_{e \in E}  h_e  = \lambda \|h\|_1.
\end{align*}
This and \eqref{upperbound-p'} imply that $\max_{\hat w \in \overline\calB_p( w, \lambda)} \sum_{e \in E} (\hat w_e -  w_e)  h_e =\lambda \|h\|_1$ with $\hat w = \hat w^*$ being a maximizer, and further notice that $\hat w^* \ge 0$. The conclusion of the Proposition for the case $p=\infty$ then follows from  this and  \eqref{translate}.
\end{proof}

\subsection{Proof for Proposition~\ref{prop:connection_global_local}}

\begin{proof}
Let define
\begin{equation}
\widetilde \calU_p(\calT, \lambda) \Let \left\{\hat \calT = \calT(\hat w) \mid \hat w \in \overline\calB_p( w, \lambda) \right\},
\end{equation}
and suppose that $\alpha = \beta = \lambda \mathds{1}$, it follows from the definition of the $\ell_\infty$-norm that $\calU(\calT, \alpha, \beta)= \widetilde \calU_\infty(\calT, \lambda)$. Thus,
\begin{equation}\label{eq:connection_global_local_tmp}
\RT_{\calU(\calT, \alpha, \beta)}(\cdot, \cdot) = \widetilde \RT_{\calU_\infty(\calT, \lambda)}(\cdot, \cdot),
\end{equation}
where we recall that $\widetilde \RT_{\calU_\infty}$ is defined in \eqref{eq:RTW_edges_lpball_woNon} with $p=\infty$.

Additionally, following the proof for Proposition~\ref{prop:globalRTW_closedform}, we also have 
\begin{equation}\label{eq:woNon}
\widetilde \RT_{\calU_\infty}(\cdot, \cdot) = \RT_{\calU_\infty}(\cdot, \cdot).
\end{equation}
Hence, we have
\begin{equation}\label{eq:connection_global_local}
\RT_{\calU(\calT, \alpha, \beta)}(\cdot, \cdot) = \RT_{\calU_\infty(\calT, \lambda)}(\cdot, \cdot).
\end{equation}
Thanks to formula \eqref{closed-form-local} for $\RT_{\calU(\calT, \alpha, \beta)}$ which is independent of $\alpha$, we can further drop the condition $\alpha = \lambda \mathds{1}$. That is, connection \eqref{eq:connection_global_local} holds true under the only condition $\beta = \lambda \mathds{1}$. Thus, the proof is completed.

\end{proof}
\subsection{Proof for Theorem~\ref{thrm:negative_definite}}

\begin{proof}
    We have $(a, b)\in \R\times\R \mapsto (a-b)^2$ is negative definite.
  
    Therefore, following \citep[Corollary~2.10, pp. 78]{Berg84}, for $1 \le p \le 2$, then we have
    \[
    (a, b) \in \R\times\R \mapsto |a-b|^p
    \]
    is negative definite. 
    
    Thus, for $1 \le p \le 2$, the mapping function
    \[
    (x, z)\in \R^d\times \R^d \mapsto \norm{x - z}_p^p
    \]
    is negative definite since it is a sum of negative definite functions. 
    
    Again, applying \citep[Corollary~2.10, pp.78]{Berg84}, we conclude that 
   \[
   (x, z) \in \R^d\times \R^d 
 \mapsto \norm{x - z}_p
 \]
  is negative definite when  $1 \le p \le 2$.

    Moreover, by using the mapping function
    \[
    \mu \mapsto \left([w_e +\beta_e]\mu(\gamma_e)\right)_{e \in E} \in \RR^{|E|}_+,
    \]
     and thanks to formula \eqref{closed-form-local}, we can reformulate $\RT_{\calU}$ between two probability measures in $\calP(V)$ in \eqref{eq:RTW_local} as the $\ell_1$ metric between two corresponding mapped vectors in $\RR^{|E|}_+$. Therefore, $\RT_{\calU}$ is negative definite.
    
    Similarly, due to formula \eqref{eq:RTW_lpball} we can also reformulate $\RT_{\calU_p}(\mu,\nu)$ 
    for $\mu,\nu\in \calP(V)$ 
    as a nonnegative weighted sum of $\ell_1$ metric (i.e., under the mapping $\mu \mapsto \left( w_e\mu(\gamma_e)\right)_{e \in E} \in \RR^{|E|}_+$) and $\ell_{p'}$ metric (i.e., under the mapping $\mu \mapsto \left(\mu(\gamma_e)\right)_{e \in E} \in \RR^{|E|}_+$) of corresponding mapped vectors in $\RR^{|E|}_+$, where $p'$ is the conjugate of $p$. In addition, $1 \le p' \le 2$ when $2\le p\le \infty$. Thus, $\RT_{\calU_p}$ is also negative definite for $2\le p\le \infty$. 
    
    Hence, the proof is complete.
\end{proof}

\subsection{Proof for Proposition~\ref{prop:divisibility}}

\begin{proof}
For probability measures $\mu, \, \nu \in \calP(V)$ and $m \in N^{*}$, we define kernel
\[
k_{\RT_{\calU}}^m(\mu, \nu) \Let 
 \exp\left(-t \frac{\RT_{\calU}(\mu, \nu)}{m}\right).
 \]
 We have $(k_{\RT_{\calU}}^m)^m(\mu, \nu) = k_{\RT_{\calU}}(\mu, \nu)$ and note that $k_{\RT_{\calU}}^m$ is a positive definite kernel. Therefore, following~\citep[\S3, Definition 2.6, pp.76]{Berg84}, kernel $k_{\RT_{\calU}}$ is infinitely divisible. 

 Similarly, for all $2\le p\le \infty$, we define kernel
 \[
 k_{\RT_{\calU_p}}^m \Let 
 \exp\left(-t \frac{\RT_{\calU_p}(\mu, \nu)}{m}\right).
 \]
 We have $(k_{\RT_{\calU_p}}^m)^m = k_{\RT_{\calU_p}}$ and note that $k_{\RT_{\calU_p}}^m$ is a positive definite kernel. Therefore, following~\citep[\S3, Definition 2.6, pp.76]{Berg84}, kernel $k_{\RT_{\calU_p}}$ is infinitely divisible. Thus, the proof is complete.
 
\end{proof}

\subsection{Proof for Proposition~\ref{prop:metric}}

\begin{proof}
    As in the proof for Theorem~\ref{thrm:negative_definite}, $\RT_{\calU}$ between two probability measures in $\calP(V)$ can be rewritten as a $\ell_1$ metric between two corresponding mapped vectors in $\RR^{|E|}_+$, i.e., by using the mapping 
    \[
    \mu \mapsto \left([w_e +\beta_e]\mu(\gamma_e)\right)_{e \in E} \in \RR^{|E|}_+.
    \]
     Therefore, $\RT_{\calU}$ is a metric.

    Similarly, $\RT_{\calU_p}$ between two probability measures in $\calP(V)$ can be recasted as a nonnegative weighted sum of $\ell_1$ metric, i.e., by the mapping 
    \[
    \mu \mapsto \left( w_e\mu(\gamma_e)\right)_{e \in E} \in \RR^{|E|}_+,
    \]
    and $\ell_{p'}$ metric, i.e., by the mapping 
    \[
    \mu \mapsto \left(\mu(\gamma_e)\right)_{e \in E} \in \RR^{|E|}_+
    \]
     of corresponding mapped vectors in $\RR^{|E|}_+$, where $p'$ is the conjugate of $p$. Thus, $\RT_{\calU_p}$ is  a metric for all $1 \le p \le \infty$. Hence, the proof is complete.
\end{proof}




\section{FURTHER DISCUSSION}\label{appsec:discussion}

\subsection{Related Work}

We give further discussion to other related works.

\paragraph{For tree-(sliced-)Wasserstein~\citep{LYFC}.} Recall that in this work, we consider OT problem for measures with noisy tree metric. In case, one uses the tree-Wasserstein (TW)~\citep{LYFC} for such problem, its performances may be affected due to the noise on the ground cost since TW fundamentally depends on the underlying tree metric structures over supports, which agrees with our empirical observations in Section~\ref{sec:experiments}.

\textbf{$\bullet$ Problem setting.} \citet{LYFC} considers the OT problem for measures with tree metric. For applications with given tree metric, one can directly apply the TW for such applications. For applications without given tree metric, \citet{LYFC} proposed to adaptively sample tree metric for supports of input measures, e.g., partition-based tree metric sampling method for supports in low-dimensional space, or clustering-based tree metric sampling method for supports potentially in high-dimensional space. Whereas in our problem, we consider measures with a given noisy tree metric. In other words, we focus on how to deal with the noise on the ground cost, and to reduce this noise affect on performances for OT problem for measures with tree metric, or TW. Although in this work, we consider a simple setting where the tree metric is given, one can leverage the clustering-based tree metric sampling measures to extend it for applications without given tree metric.

\textbf{$\bullet$ Extension for general applications via tree metric sampling.} For applications without given tree metric, but with Euclidean supports, one can apply the clustering-based tree metric sampling method~\citep{LYFC} to obtain a tree metric for such applications. Such sampled tree metric may be noisy, e.g., due to perturbation on supports of input measures, or noisy/adversarial measurement within the clustering-based tree metric sampling method (i.e., clustering algorithm or initialization). Thus, our approach can tackle for such problems in applications. 

In our work, to deal with the noise on tree metric for OT problem, we follow the max-min robust OT for measures with noisy tree metric. As illustrated in our experiments in Section~\ref{sec:experiments}, the robust OT approach improves performances of the counterpart standard OT when ground metric cost is noisy.

\textbf{$\bullet$ Potential combination.} There is a potential to combine our approach with the approach in~\citep{LYFC} together. For example, when the tree metric space for supports of probability measures is not given, and we can query tree metrics from an oracle, but only receive perturbed/noisy tree metrics. It is an interesting direction for further investigation.

We further note that averaging over TW corresponding to those sampled tree metrics in~\citep{LYFC} has the benefit of reducing clustering/quantization sensitivity problems for the tree metric sampling methods in~\citep{LYFC}.

\paragraph{For $1$-Wasserstein approximation with TW~\citep{yamada2022approximating}.} \citet{yamada2022approximating} considered to use TW as an approximation model for some given (oracle) $1$-Wasserstein distance. In particular, given some supervised triplets $(\mu, \nu, \calW(\mu, \nu))$ with measures $\mu, \nu$ being supported on high-dimensional vector space $\RR^n$. The ground cost metric for the observed $1$-Wasserstein is unknown. \citet{yamada2022approximating} used TW as a model to fit the observed $1$-Wasserstein distances, i.e., estimate tree metric such that $\calW_{\calT}(\cdot, \cdot) = \calW(\cdot, \cdot)$. Therefore, the approach in \citet{yamada2022approximating} cannot be used in our considered problem due to the following two main reasons:
\begin{itemize}
    \item (i) we do not have such supervised clean $1$-Wasserstein distance data in our considered problem (i.e., the observed $1$-Wasserstein distance $\calW(\mu, \nu)$ corresponding with two input probability measures $\mu$ and $\nu$).

    \item (ii) our considered probability measures are supported in a given tree metric space $(\calT, d_{\calT})$. Moreover, the given tree structure is not necessarily a physical tree in the sense that the set of vertices $V$ is a subset of some high-dimensional vector space $\RR^n$, and each edge $e$ is the standard line segment in $\RR^n$ connecting the two end-points of edge $e$. The given tree structure $\calT$ in our problem can be non-physical, while in \citet{yamada2022approximating}, input probability measures $\mu, \nu$ are supported on some high-dimensional vector space $\R^n$.
\end{itemize}

\subsection{Other Discussions}

\paragraph{For kernels on probability measures with OT geometry.} In general, Wasserstein space is \emph{not} Hilbertian~\citep[\S8.3]{peyre2019computational}, and the standard OT is indefinite. Thus, it is nontrivial to build positive definite kernels upon OT for probability measures.

For kernels on probability measures with OT geometry, besides the tree-(sliced)-Wasserstein kernel~\citep{LYFC}, and sliced-Wasserstein kernel~\citep{kolouri2016sliced, Carriere-2017-Sliced}, to our knowledge, there are only the permanent kernel~\citep{cuturi2007kernel} and generating function kernel~\citep{cuturi2012positivity}. However, they are intractable.

For kernels on general measures with potentially different total mass with OT geometry, to our knowledge, there is only the entropy partial transport kernel~\citep{le2021ept}.

\paragraph{For subspace robust Wasserstein~\citep{pmlr-v97-paty19a}.} In this work, we considered OT problem for probability measures with noisy tree metric. Recall that the subspace robust Wasserstein (SRW)~\citep{pmlr-v97-paty19a}  assumes probability measures supported on high-dimensional vector space $\RR^n$ and seeks the robustness over its vector subspaces (i.e., $\R^m$ with $m < n$). Therefore, the SRW distance is not applicable for the considered problem since it is not clear what is the reasonable notion of subspaces of a given tree metric space in our considered problem. Additionally, note that the given tree structure $\calT$ in our problem is not necessarily physical, i.e., it can be a non-physical tree structure.

Moreover, for measures supported on high-dimensional Euclidean space (e.g., $\RR^{300}$), it takes too much time for the computation of subspace robust Wasserstein (SRW), and the SRW does not scale up for large-scale settings (e.g., see Section~\ref{appsubsec:further_experiments}).

\paragraph{For noise in the document datasets.} In our empirical simulations, we emphasize that our purposes are to compare different transport approaches for probability measures supported on a given tree metric space in the \emph{same} settings. 

Moreover, as discussed in~\citet{sato2022re}, the duplication is not the problem for comparing performances in noisy environments. Indeed, in our simulations, we keep the same settings, the only difference is the transport distance. We have no assumptions/requirements that the considered document datasets are clean in our simulations. Therefore, the noise in the considered document datasets are not a problem for our simulations, but provides more diversity for settings in our simulations.

\paragraph{For tree metric.} We applied the clustering-based tree metric sampling method in~\citet{LYFC} to obtain a tree metric and use it as the original given tree metric without perturbation. We emphasize that there is no further processing on that tree metric without perturbation.

For simulations for measures with noisy tree metric, as detailed in Section \ref{sec:experiments} in the main manuscript, we specifically generate perturbations on each edge length of the original given tree by deviating it by a random nonnegative amount which is less than or equal $\Delta \in \RR_+$, i.e., $|w_e - w_e^*| \le \Delta$ where $w_e^*, w_e$ are edge lengths on the original tree without perturbation and on the given perturbed tree respectively. We also emphasize that for all edge $e$ in the perturbed tree, we preserve the condition $w_e \ge 0$. When there exist edge $e$ with $0$-length, i.e., $w_e = 0$ in the noisy tree, it can be interpreted that the perturbed tree not only changes its edge lengths, but also its structure (see Theorem~\ref{thrm:zeroedges}).

Note that, when there is no perturbation on the given tree metric, it is equivalent to $\Delta = 0$.

\paragraph{For time consumptions of the robust OT.} As in \S\ref{sec:RobustTW} in the main manuscript, we show that the computational complexity of the max-min robust OT for measures with tree metric is linear to the number of edges in tree $\calT$ which is in the same order as the counterpart TW, i.e., standard OT with tree metric ground cost. Moreover, in general, the max-min robust OT is hard and expensive to compute due to its non-convexity and non-smoothness, i.e., a maximization problem w.r.t. tree metric with OT as its objective function. This problem remains hard even for supports in $1$-dimensional space~\citep{deshpande2019max}. We further note that even with a given optimal ground metric cost, the computational complexity of max-min/min-max robust OT is in the same order as their counterpart standard OT (i.e., their objective function).

As in the experimental results in Section~\ref{sec:experiments}, illustrated in Figures \ref{fg:DOC_Noise_Delta0.5}, and \ref{fg:TDA_Noise_Delta0.05}, the time consumptions of the robust OT are comparable with the counterpart TW (i.e., OT with tree metric ground cost) for measures with noisy tree metric. The robust OT with the global approach is a little slower than that of others. This is a stark contrast to other approaches for max-min robust OT problem in general. The proposed novel uncertainty sets of tree metrics play an important role for scalability of robust OT for measures with tree metric (e.g., comparing to the one-dimensional OT-based approach for max-min robust OT~\citep{deshpande2019max} where there is no efficient/fast algorithmic approach for it yet.)

\paragraph{For performances of the robust OT.} We emphasize that we consider the OT problem for measures with noisy tree metric. The empirical results in Figures~\ref{fg:DOC_Noise_Delta0.5}, and \ref{fg:TDA_Noise_Delta0.05} illustrate that the max-min robust OT approach helps to mitigate this issue in applications. When the given tree metric is perturbed, the performances of the proposed kernels $k_{\RT_{\calU}}$ and $k_{\RT_{\calU_2}}$ compare favorably to those of the counterpart standard OT (i.e., TW) kernel $k_{\TW}$. 

\begin{figure*}
  \begin{center}
    \includegraphics[width=0.7\textwidth]{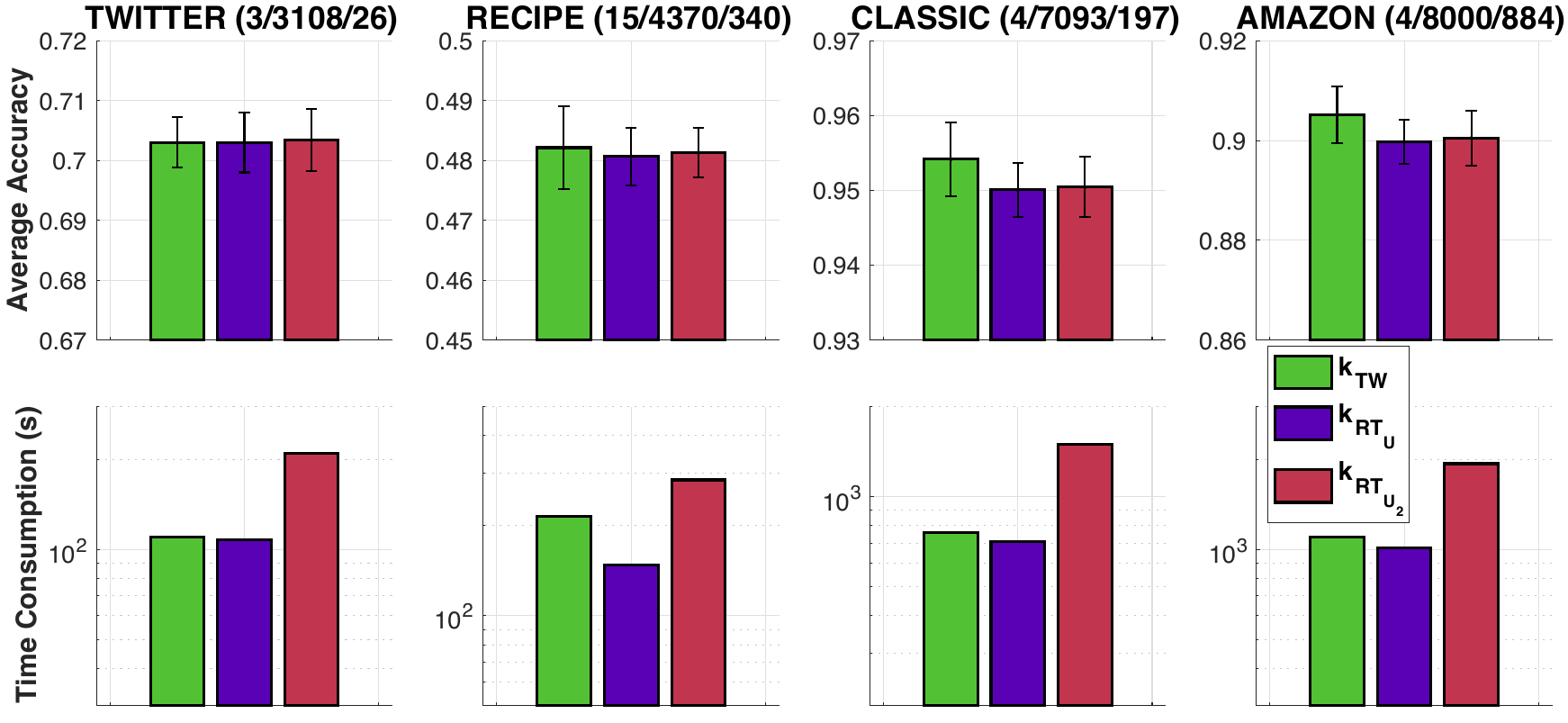}
  \end{center}
  \vspace{-6pt}
  \caption{SVM results and time consumption for kernel matrices in document classification when there is no perturbation on the given tree metric (i.e., tree metric obtained from the sampling method without perturbation) (or with $\Delta=0$). For each dataset, the numbers in the parenthesis are the number of classes; the number of documents; and the maximum number of unique words for each document respectively.}
  \label{fg:DOC_NoPerturbation}
\end{figure*}

\begin{figure}
  \begin{center}
\includegraphics[width=0.4\textwidth]{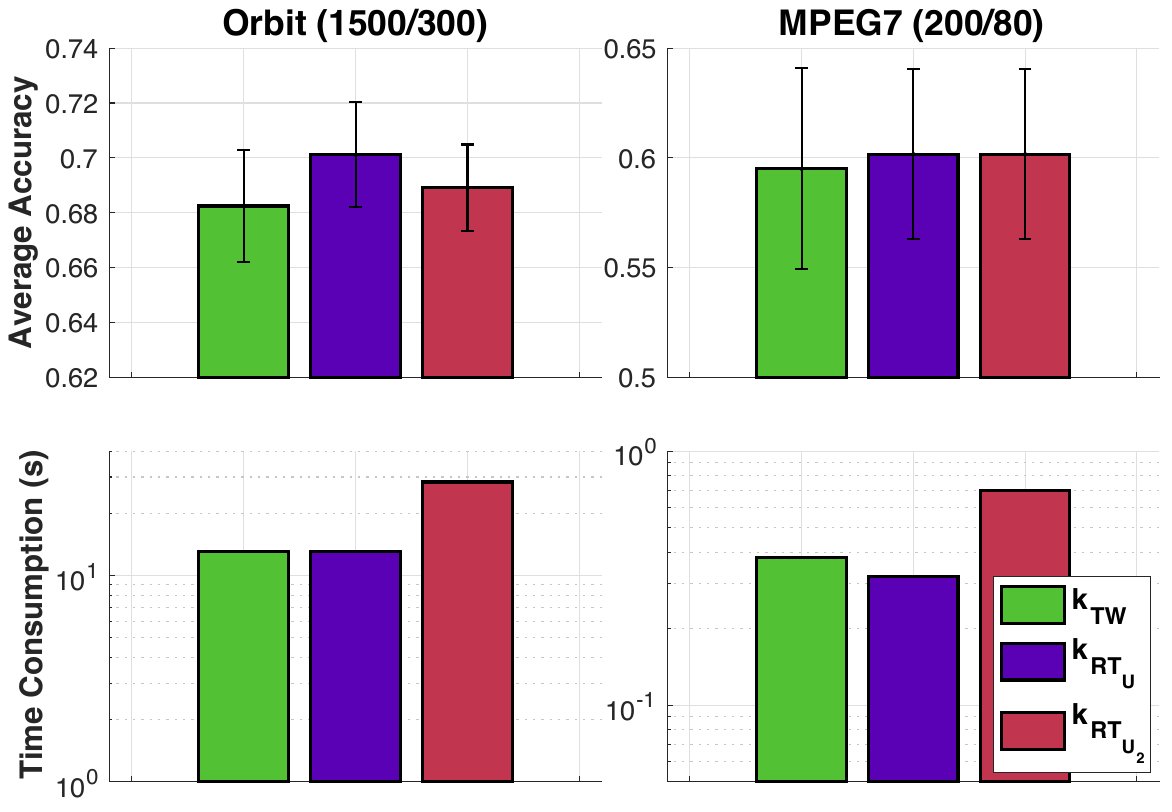}
  \end{center}
  \vspace{-6pt}
  \caption{SVM results and time consumption for kernel matrices in TDA when there is no perturbation on the given tree metric (i.e., tree metric obtained from the sampling method without perturbation) (or with $\Delta=0$). For each dataset, the numbers in the parenthesis are respectively the number of PD; and the maximum number of points in PD.}
  \label{fg:TDA_NoPerturbation}
\end{figure}

Additionally, Figures~\ref{fg:DOC_NoPerturbation}, \ref{fg:TDA_NoPerturbation} illustrate further empirical results where the given tree metric is directly obtained from the sampling method without perturbation (or with $\Delta=0$).\footnote{We have not argued advantages of the max-min robust OT over the counterpart standard OT for such problems.}. The performances of the proposed kernels for the max-min robust OT $k_{\RT_{\calU}}$ and $k_{\RT_{\calU_2}}$ are comparable to the counterpart standard OT (i.e., TW) kernel $k_{\TW}$. Interestingly, in \texttt{Orbit} dataset, our proposed kernels $k_{\RT_{\calU}}$ and $k_{\RT_{\calU_2}}$ improves performances of kernel $k_{\TW}$. This may suggest that the given tree $\calT$ in \texttt{Orbit} dataset might be subjected to noise in our simulations.

Comparing SVM results in Figures~\ref{fg:DOC_NoPerturbation}, \ref{fg:TDA_NoPerturbation} for measures with original tree metric (i.e., directly obtained from the sampling method without perturbation, or with $\Delta=0$) and SVM results in Figures \ref{fg:DOC_Noise_Delta0.5}, and \ref{fg:TDA_Noise_Delta0.05} for measures with noisy tree metric (i.e., $\Delta = 0.5$), the noise on tree metric did harm performances of TW for all datasets in document classification and TDA. The robust OT approach helps to mitigate the effect of noisy metric for measures in applications.

\paragraph{Remarks.} After the submission of the main manuscript, a concurrent work was published in ArXiv~\citep{yamada2023empirical} where~\citet{yamada2023empirical} leverage the min-max robust variant of tree-Wasserstein for simplicial representation learning by employing self-supervised learning approach based on SimCLR. Empirically, \citet{yamada2023empirical} also illustrated the advantages of their proposed method over standard SimCLR and cosine-based representation learning.

\subsection{More Details about Experiments}

We describe further details about softwares and datasets.

\paragraph{Softwares.}

\begin{itemize}

\item For our simulations in TDA, we used DIPHA toolbox to extract persistence diagrams. The DIPHA toolbox is available at \url{https://github.com/DIPHA/dipha}.

\item For the clustering-based tree metric sampling, we used the MATLAB code at \url{https://github.com/lttam/TreeWasserstein}. We directly used this code for clustering-based tree metric sampling without any further processing to obtain the original tree metric without perturbation in our simulations.

\end{itemize}

\paragraph{Datasets.}

\begin{itemize}

\item For the document datasets (e.g., \texttt{TWITTER, RECIPE, CLASSIC, AMAZON}), they are available at \url{https://github.com/mkusner/wmd}.

\item For \texttt{Orbit} dataset, we follow the procedure in \cite{adams2017persistence} to generate it.

\item For \texttt{MPEG7} dataset, it is available at \url{http://www.imageprocessingplace.com/downloads_V3/root_downloads/image_databases/MPEG7_CE-Shape-1_Part_B.zip}. We then extract the $10$-class subset of the dataset as in~\citet{le2018persistence}.

\end{itemize}

\paragraph{Hyperparamter validation.} For validation, we further randomly split \emph{the training set} into $70\%/30\%$ for validation-training and validation with $10$ repeats to choose the hyperparamters in our experiments.

\section{FURTHER EXPERIMENTAL RESULTS}\label{appsec:experiment}

In this section, we give further detailed results for our simulations.

\subsection{Document Classification}

\begin{figure*}
 \begin{center}
   \includegraphics[width=0.7\textwidth]{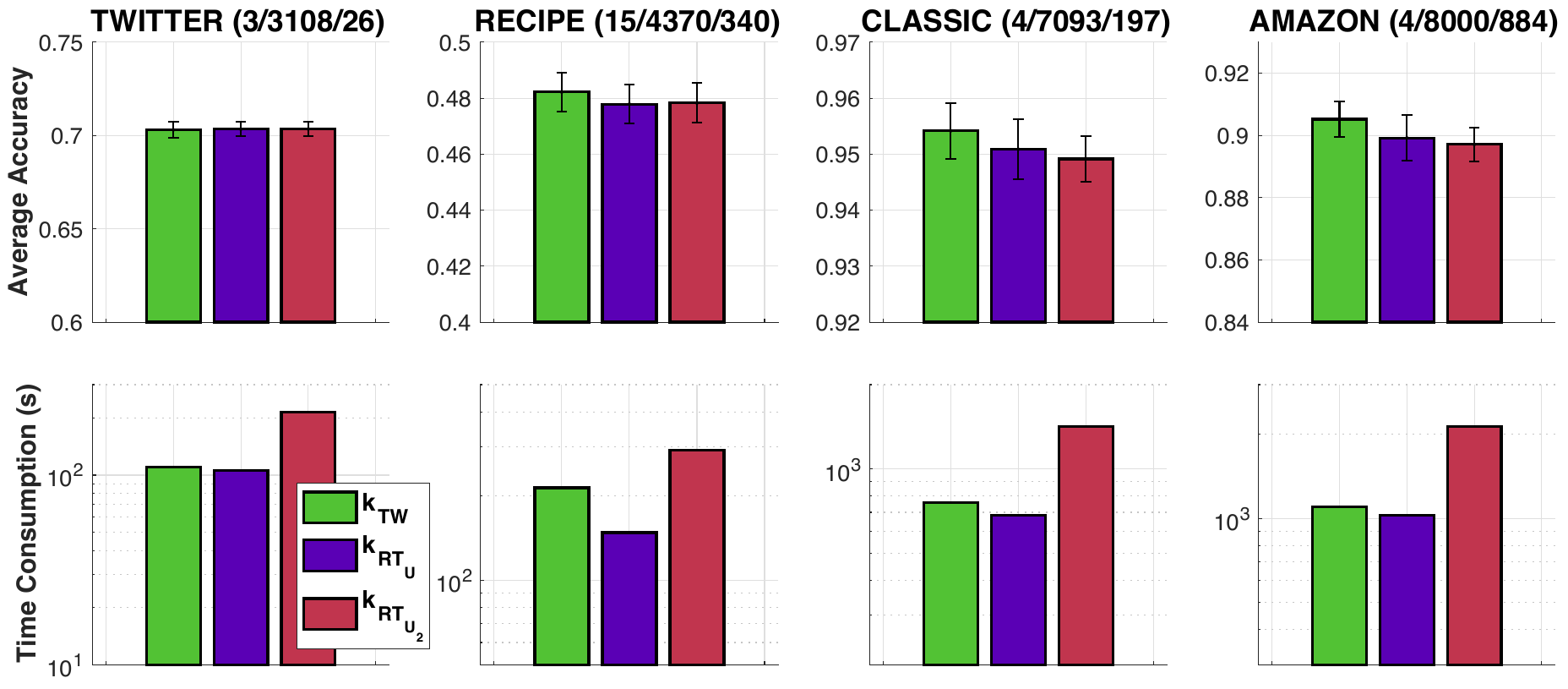}
 \end{center}
 \caption{Results on document classification for $\Delta=0$ and $\lambda = 0.01$.}
 \label{fg:DOC_NoPerturbation_A001}
\end{figure*}

\begin{figure*}
 \begin{center}
   \includegraphics[width=0.7\textwidth]{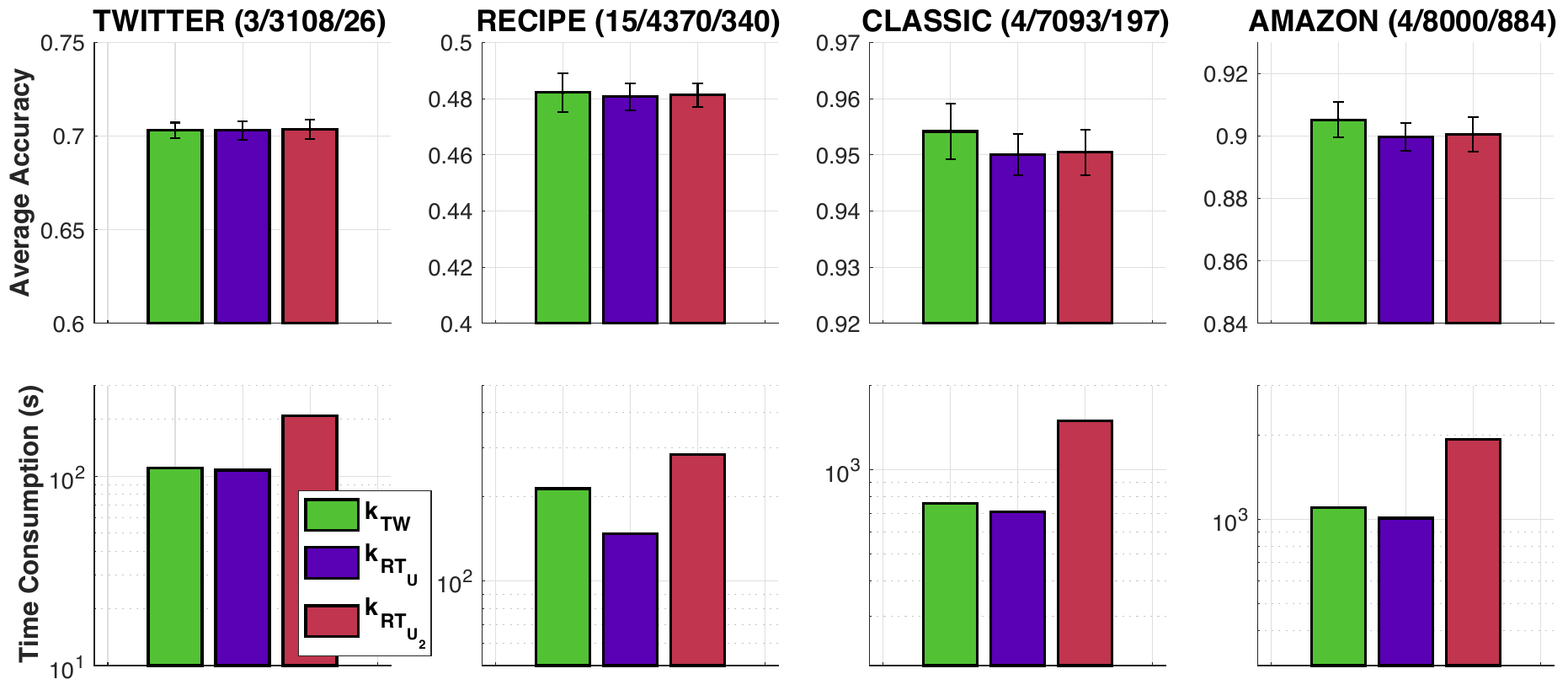}
 \end{center}
 \caption{Results on document classification for $\Delta=0$ and $\lambda = 0.05$.}
 \label{fg:DOC_NoPerturbation_A005}
\end{figure*}

\begin{figure*}
 \begin{center}
   \includegraphics[width=0.7\textwidth]{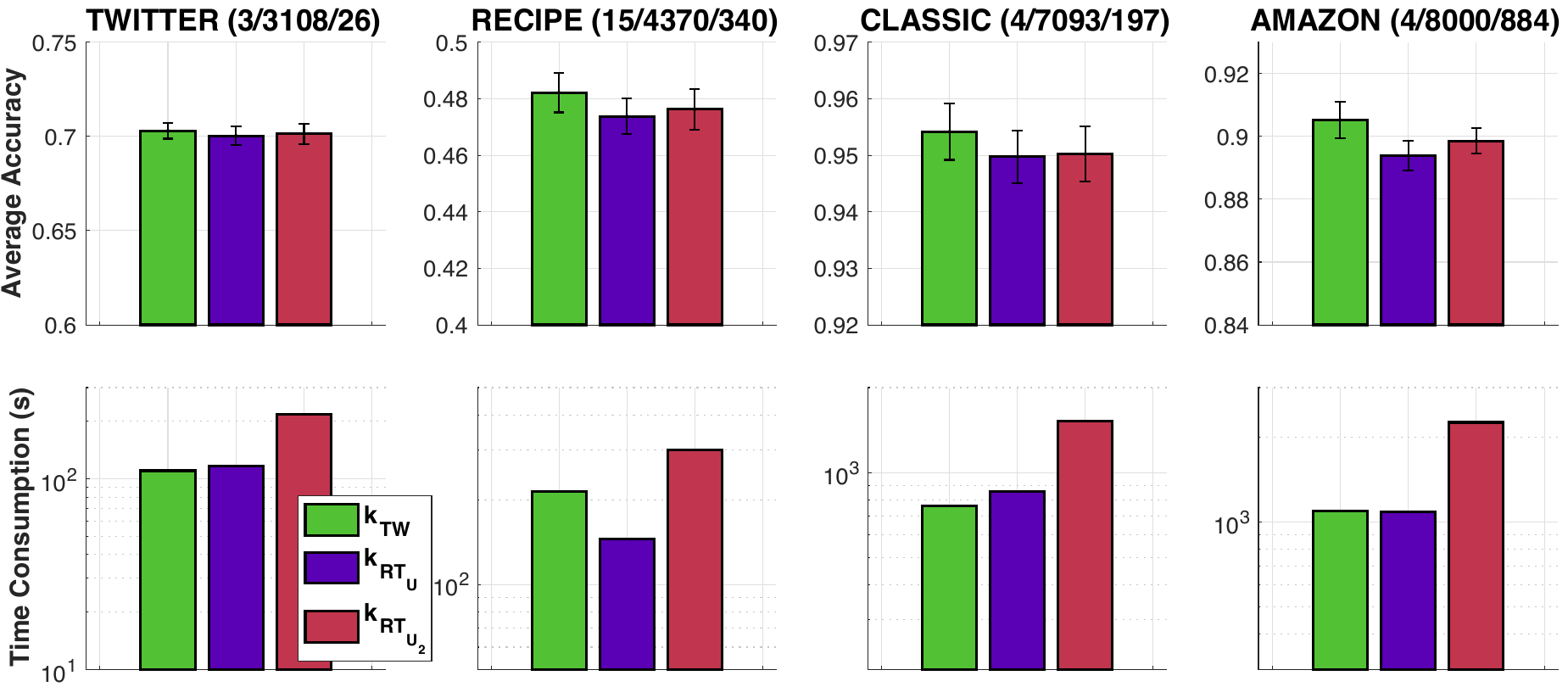}
 \end{center}
 \caption{Results on document classification for $\Delta=0$ and $\lambda = 0.1$.}
 \label{fg:DOC_NoPerturbation_A01}
\end{figure*}

\begin{figure*}
 \begin{center}
   \includegraphics[width=0.7\textwidth]{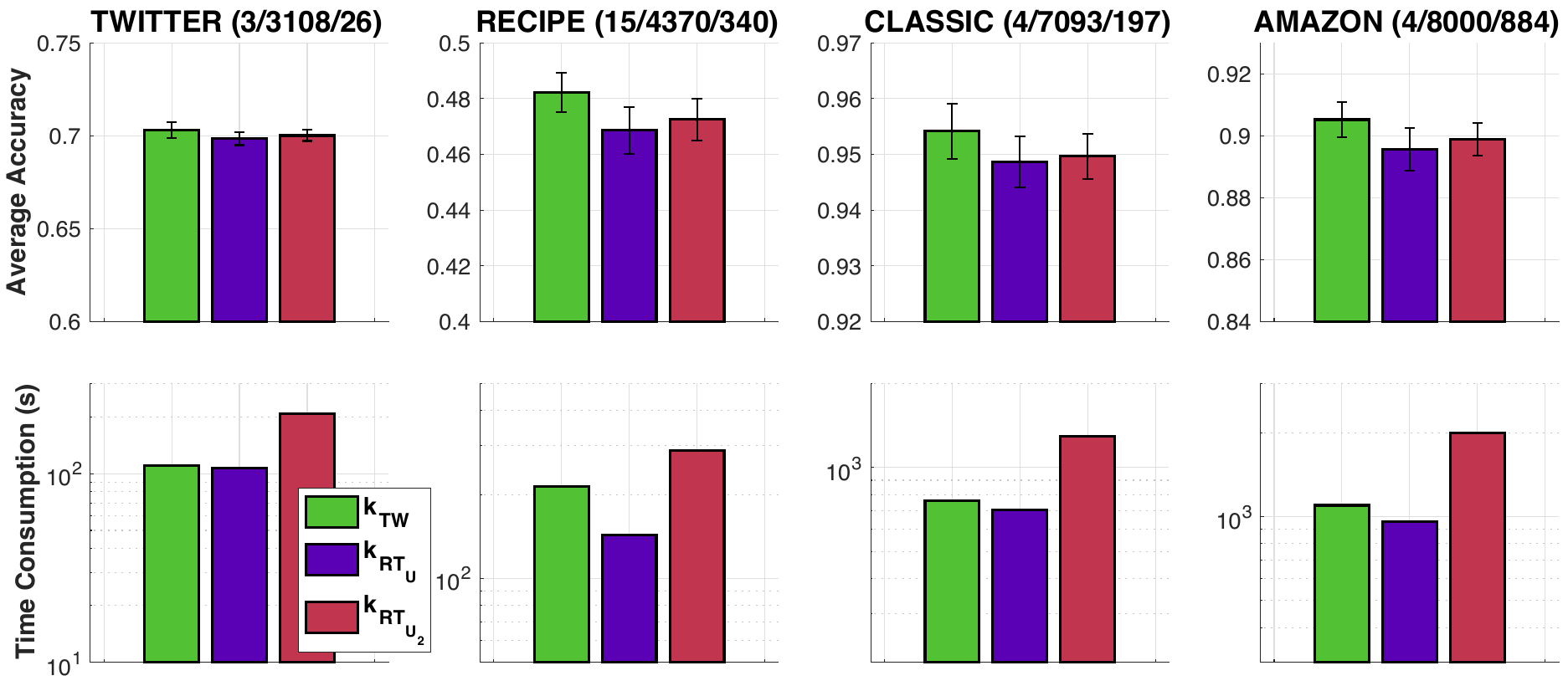}
 \end{center}
 \caption{Results on document classification for $\Delta=0$ and $\lambda = 0.5$.}
 \label{fg:DOC_NoPerturbation_A05}
\end{figure*}

\begin{figure*}
 \begin{center}
   \includegraphics[width=0.7\textwidth]{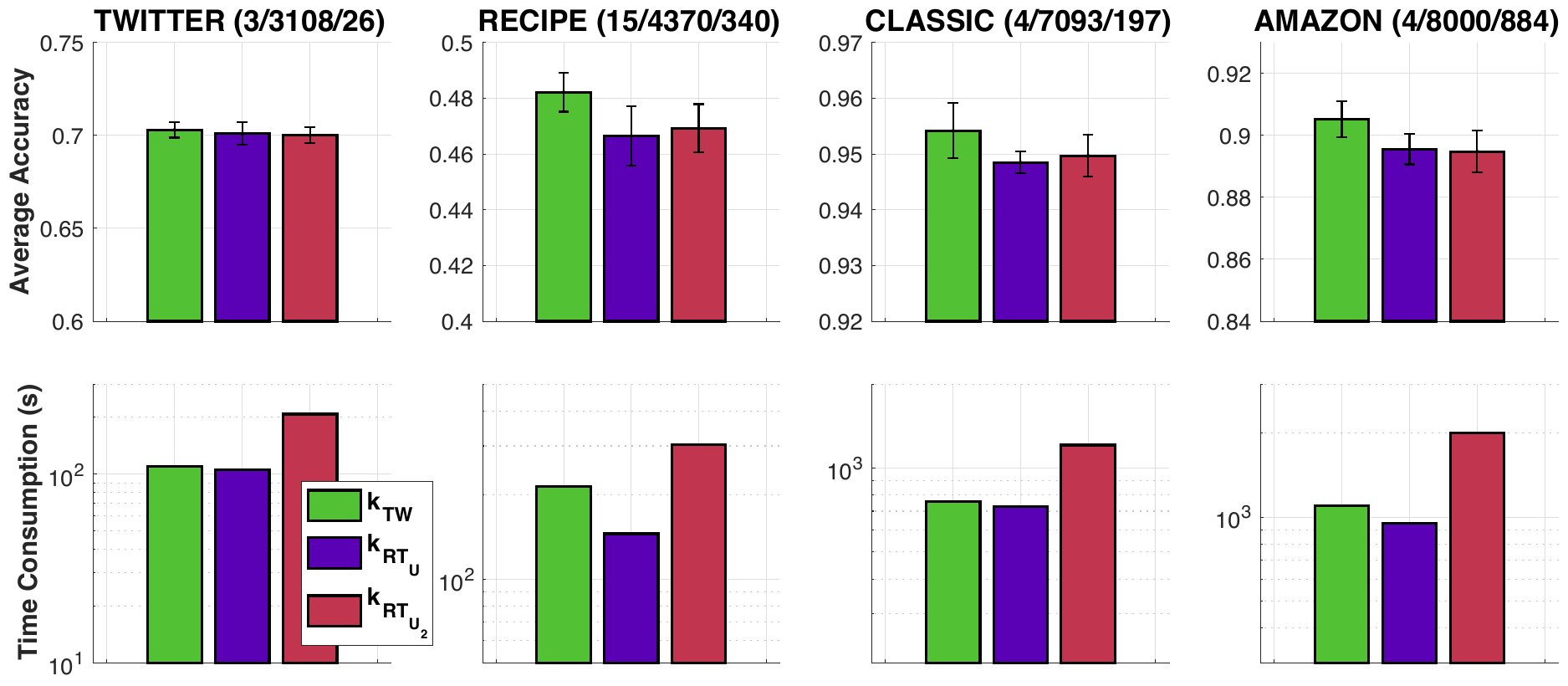}
 \end{center}
 \caption{Results on document classification for $\Delta=0$ and $\lambda = 1$.}
 \label{fg:DOC_NoPerturbation_A1}
\end{figure*}

\begin{figure*}
 \begin{center}
   \includegraphics[width=0.7\textwidth]{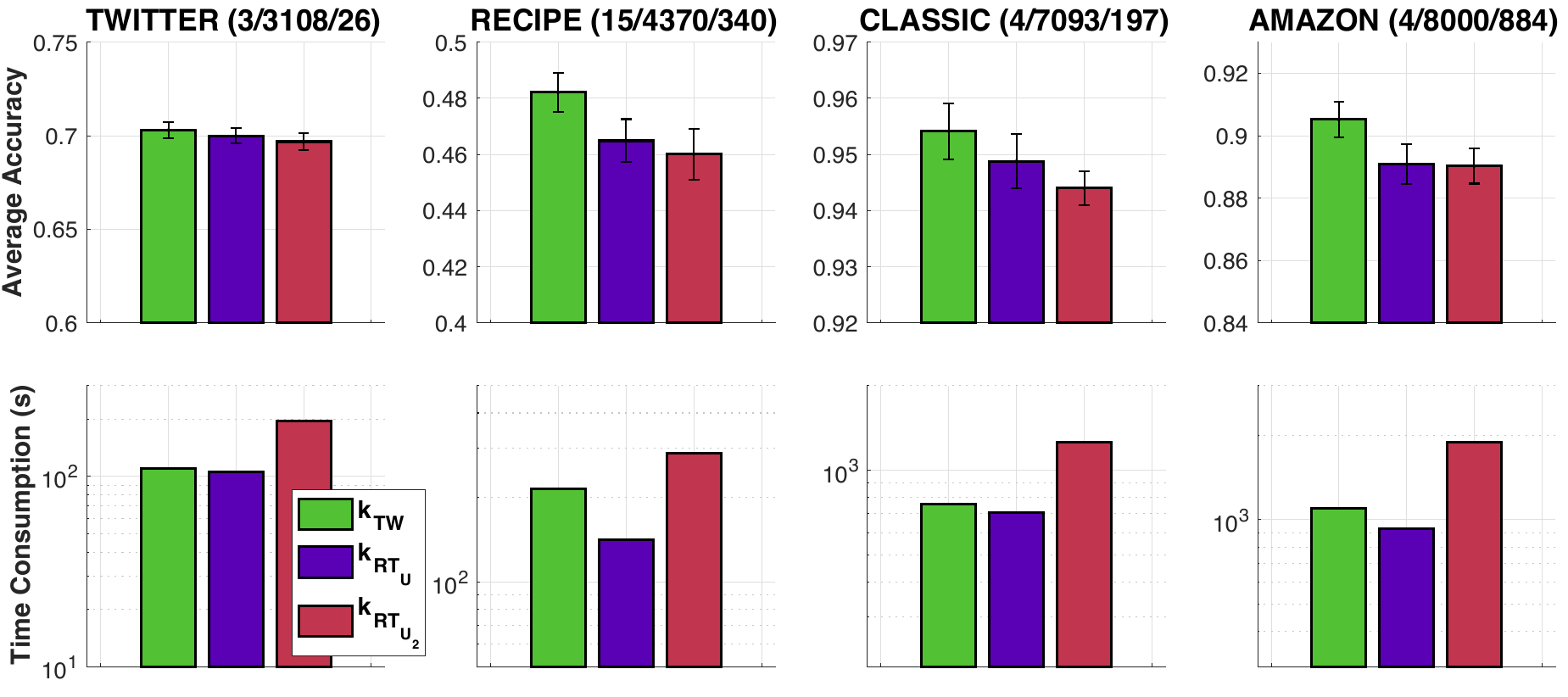}
 \end{center}
 \caption{Results on document classification for $\Delta=0$ and $\lambda = 5$.}
 \label{fg:DOC_NoPerturbation_A5}
\end{figure*}

\paragraph{For tree metric without perturbation (i.e., $\Delta = 0$).} We give detailed results for robust OT with different value of $\lambda$ in Figures~\ref{fg:DOC_NoPerturbation_A001}, \ref{fg:DOC_NoPerturbation_A005}, \ref{fg:DOC_NoPerturbation_A01}, 
\ref{fg:DOC_NoPerturbation_A05}, \ref{fg:DOC_NoPerturbation_A1}, and \ref{fg:DOC_NoPerturbation_A5}.


\begin{figure*}
 \begin{center}
   \includegraphics[width=0.7\textwidth]{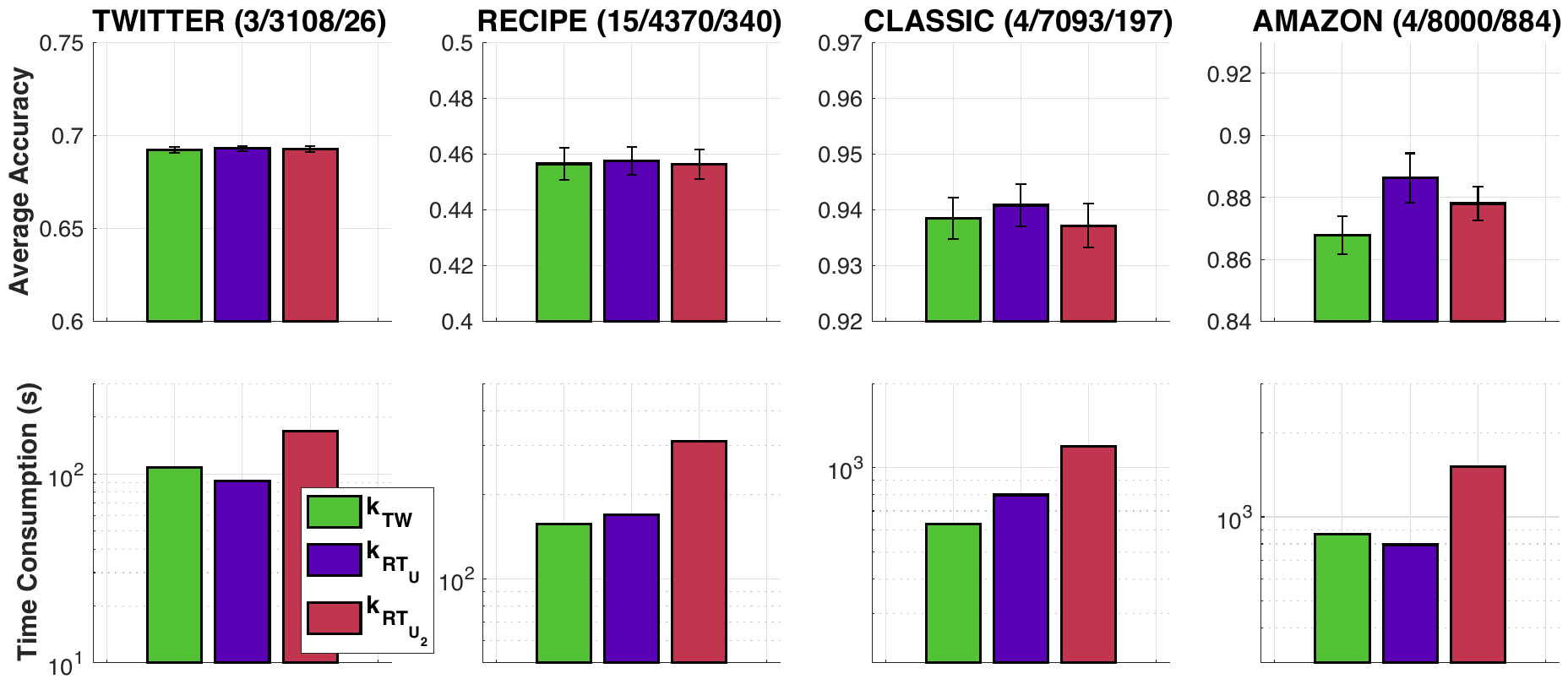}
 \end{center}
 \caption{Results on document classification for $\Delta=0.5$ and $\lambda = 0.01$.}
 \label{fg:DOC_Noise_Delta0.5_A001}
\end{figure*}

\begin{figure*}
 \begin{center}
   \includegraphics[width=0.7\textwidth]{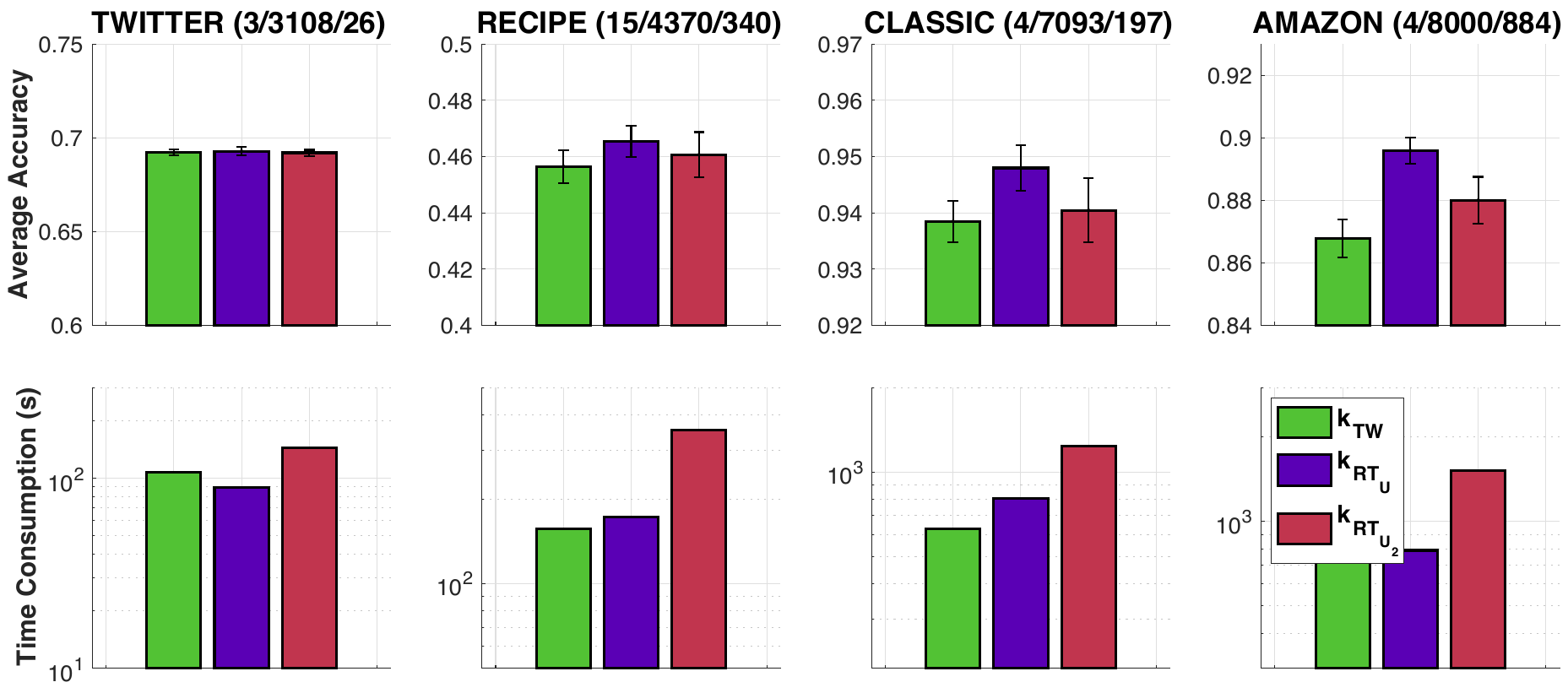}
 \end{center}
 \caption{Results on document classification for $\Delta=0.5$ and $\lambda = 0.05$.}
 \label{fg:DOC_Noise_Delta0.5_A005}
\end{figure*}

\begin{figure*}
 \begin{center}
   \includegraphics[width=0.7\textwidth]{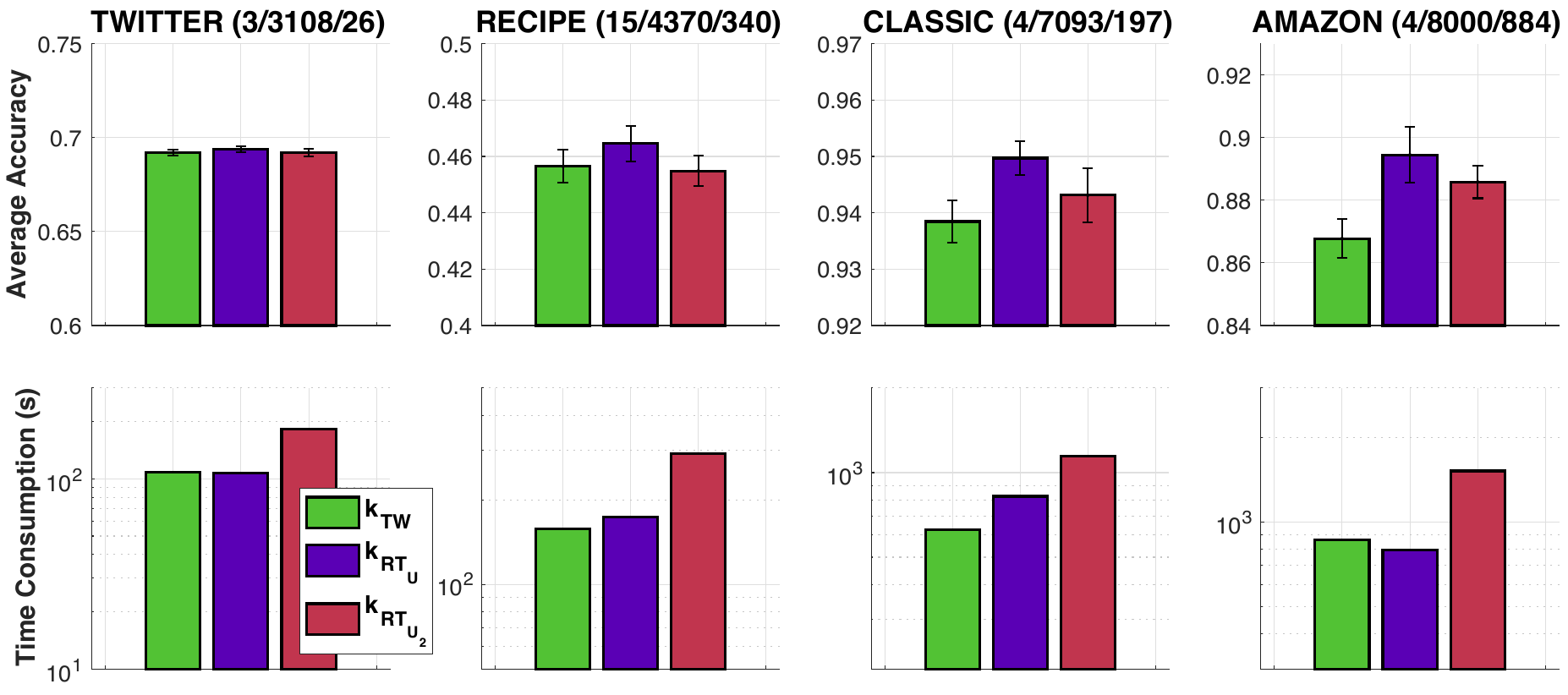}
 \end{center}
 \caption{Results on document classification for $\Delta=0.5$ and $\lambda = 0.1$.}
 \label{fg:DOC_Noise_Delta0.5_A01}
\end{figure*}

\begin{figure*}
 \begin{center}
   \includegraphics[width=0.7\textwidth]{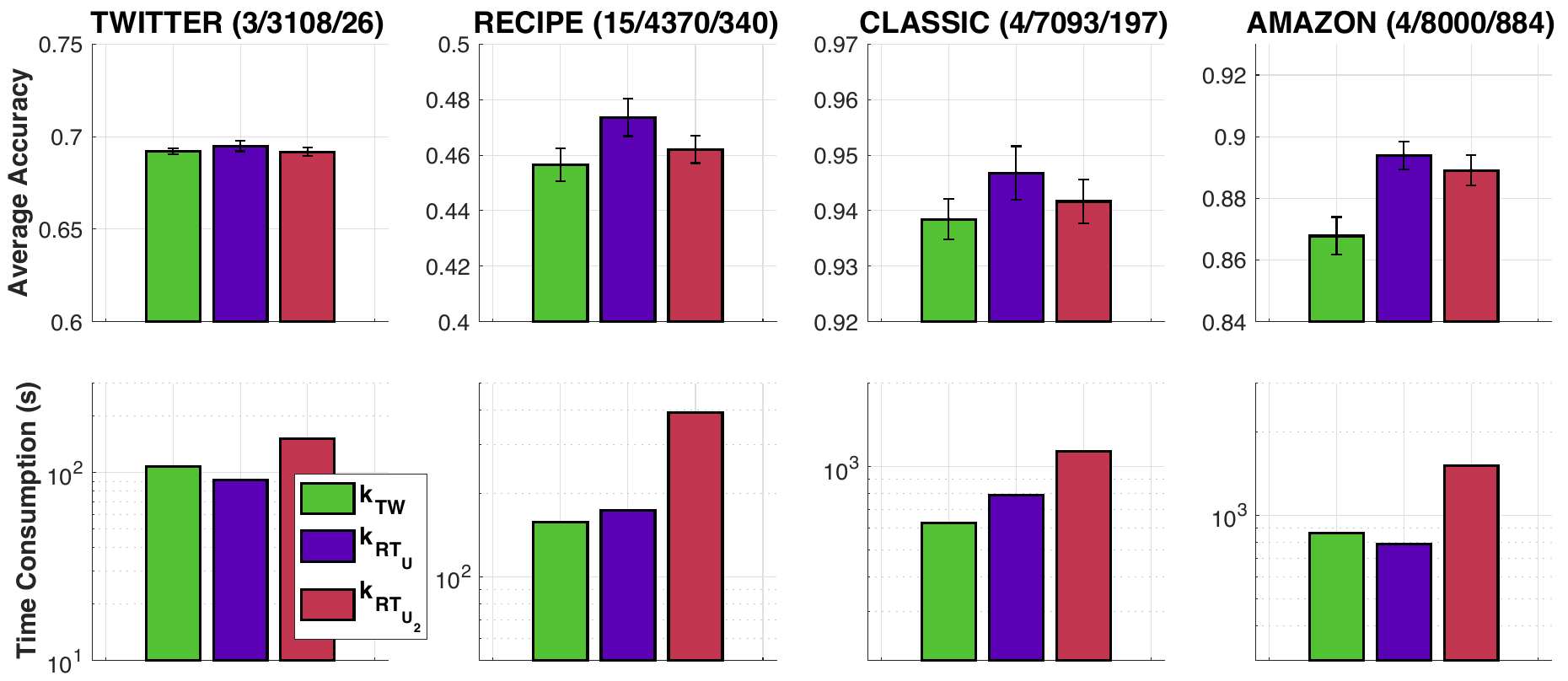}
 \end{center}
 \caption{Results on document classification for $\Delta=0.5$ and $\lambda = 0.5$.}
 \label{fg:DOC_Noise_Delta0.5_A05}
\end{figure*}

\begin{figure*}
 \begin{center}
   \includegraphics[width=0.7\textwidth]{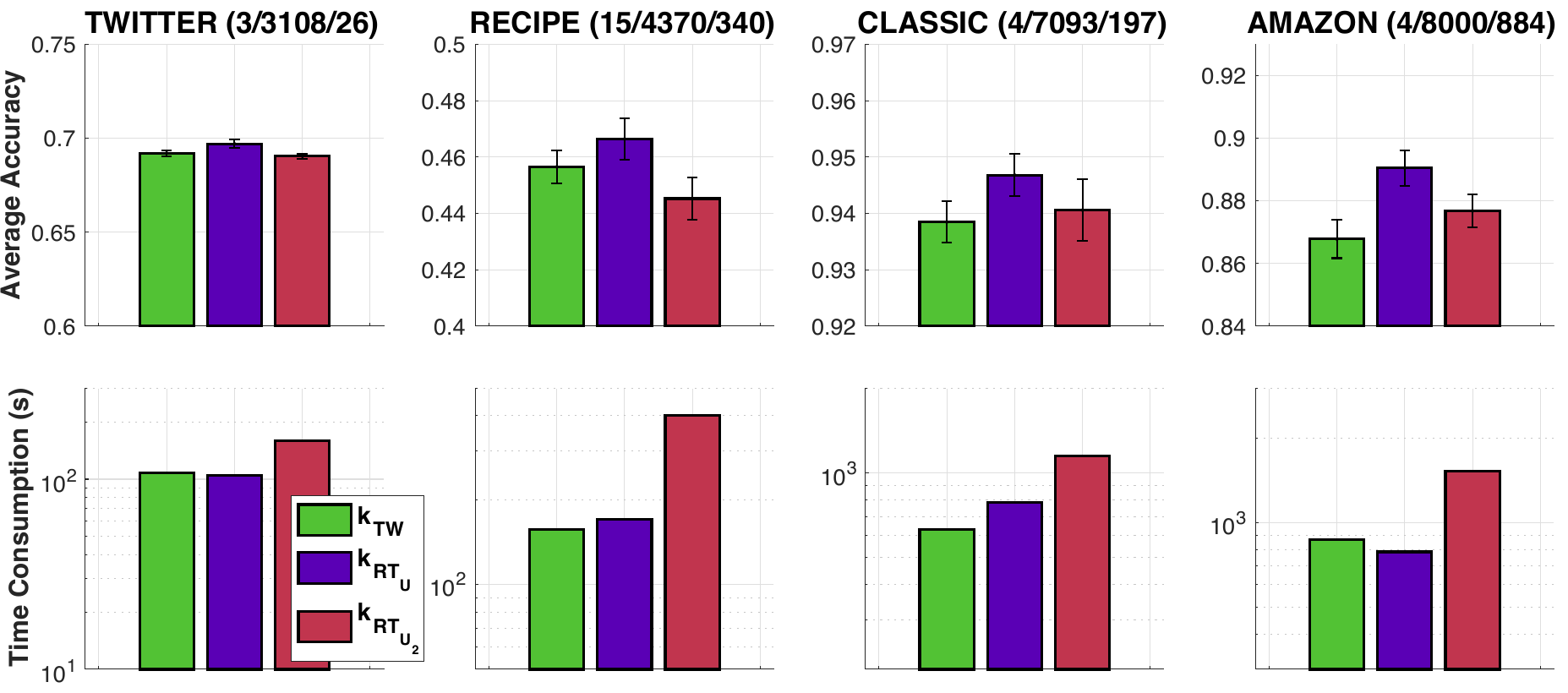}
 \end{center}
 \caption{Results on document classification for $\Delta=0.5$ and $\lambda = 1$.}
 \label{fg:DOC_Noise_Delta0.5_A1}
\end{figure*}

\begin{figure*}
 \begin{center}
   \includegraphics[width=0.7\textwidth]{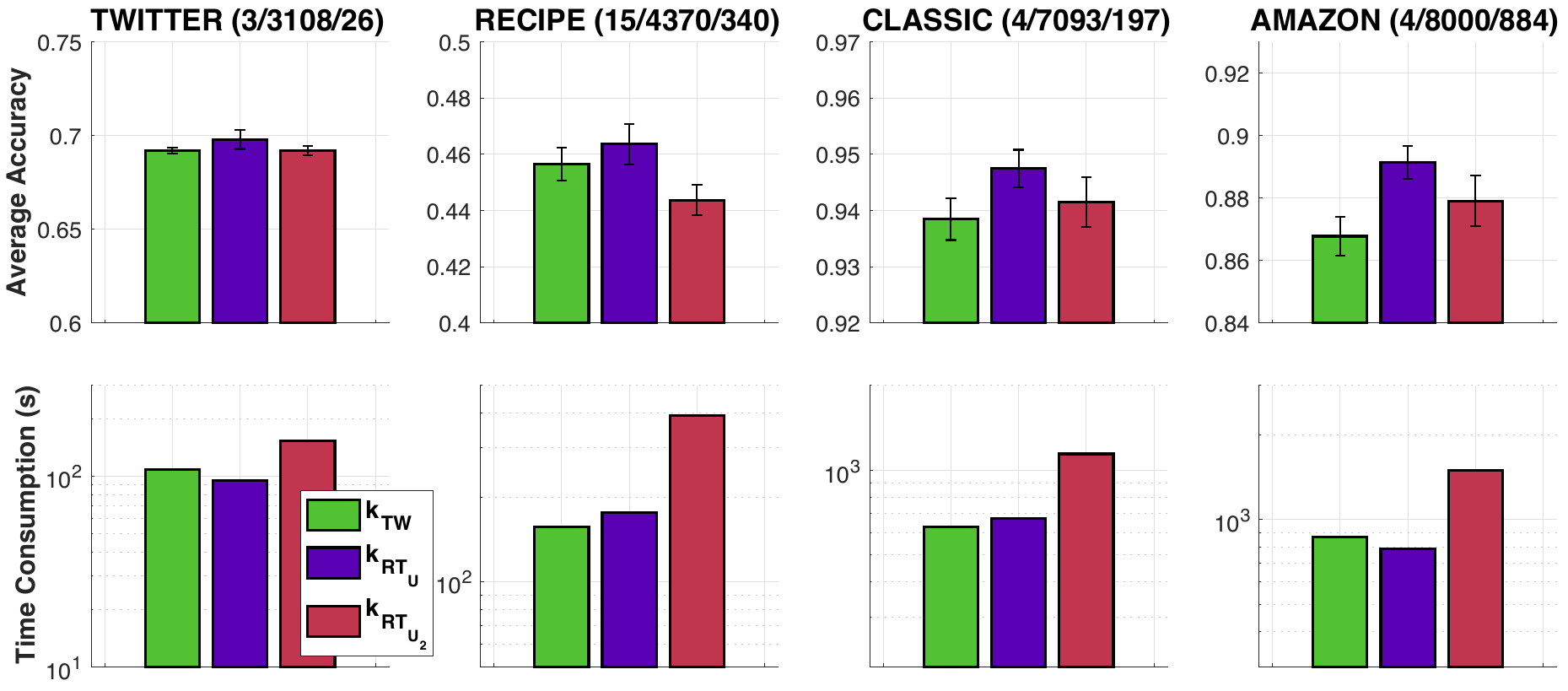}
 \end{center}
 \caption{Results on document classification for $\Delta=0.5$ and $\lambda = 5$.}
 \label{fg:DOC_Noise_Delta0.5_A5}
\end{figure*}

\paragraph{For noisy tree metric.} We give detailed results for robust OT with different value of $\lambda$ in Figures~\ref{fg:DOC_Noise_Delta0.5_A001}, \ref{fg:DOC_Noise_Delta0.5_A005}, \ref{fg:DOC_Noise_Delta0.5_A01}, 
\ref{fg:DOC_Noise_Delta0.5_A05}, \ref{fg:DOC_Noise_Delta0.5_A1}, and \ref{fg:DOC_Noise_Delta0.5_A5} when tree metric is perturbed with $\Delta = 0.5$.


\subsection{Topological Data Analysis}

\begin{figure}[h]
 \begin{center}
   \includegraphics[width=0.4\textwidth]{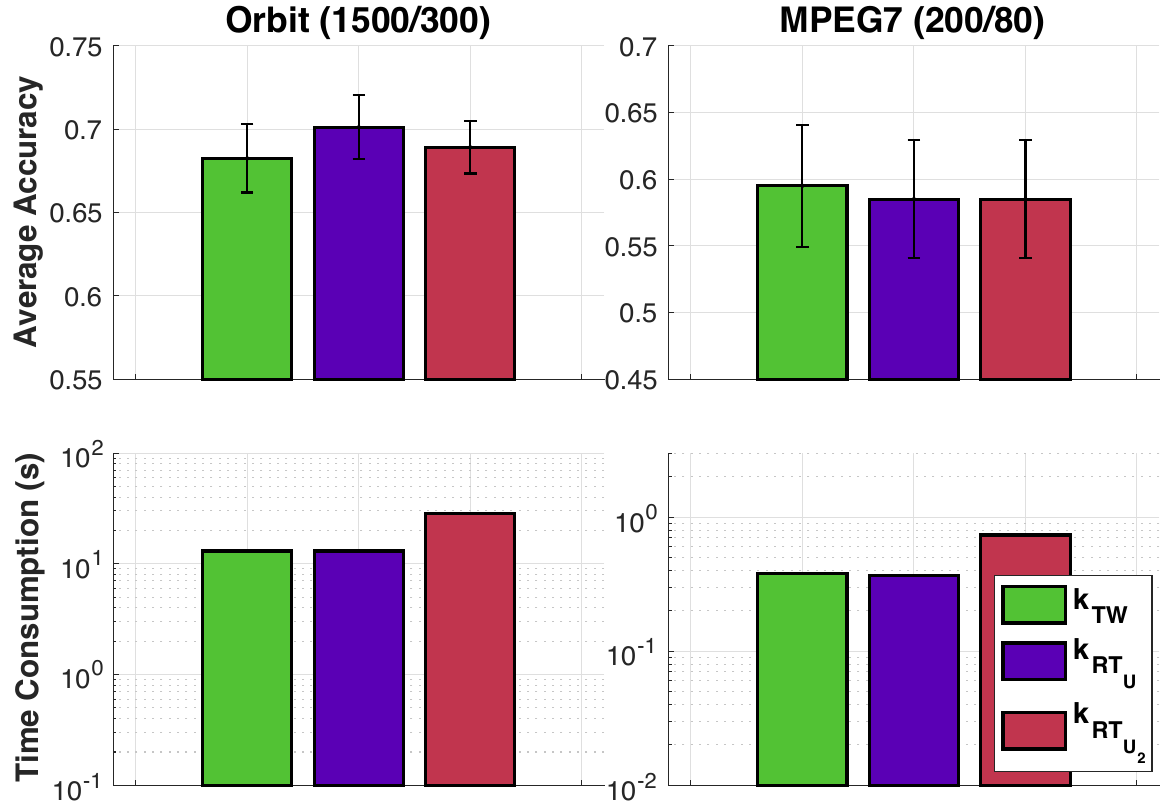}
 \end{center}
 \caption{Results on TDA for $\Delta = 0$ and $\lambda = 0.01$.}
 \label{fg:TDA_NoPerturbation_A001}
\end{figure}

\begin{figure}[h]
 \begin{center}
   \includegraphics[width=0.4\textwidth]{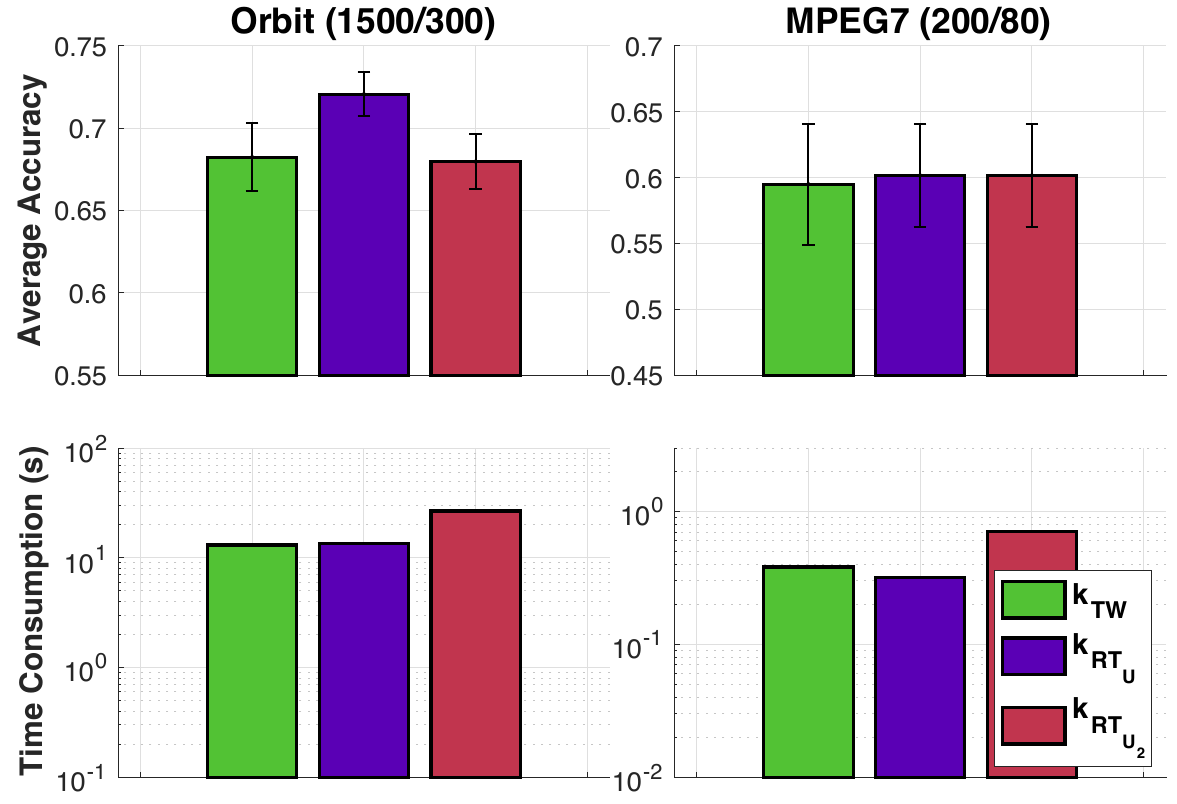}
 \end{center}
 \caption{Results on TDA for $\Delta = 0$ and $\lambda = 0.05$.}
 \label{fg:TDA_NoPerturbation_A005}
\end{figure}

\begin{figure}[h]
 \begin{center}
   \includegraphics[width=0.4\textwidth]{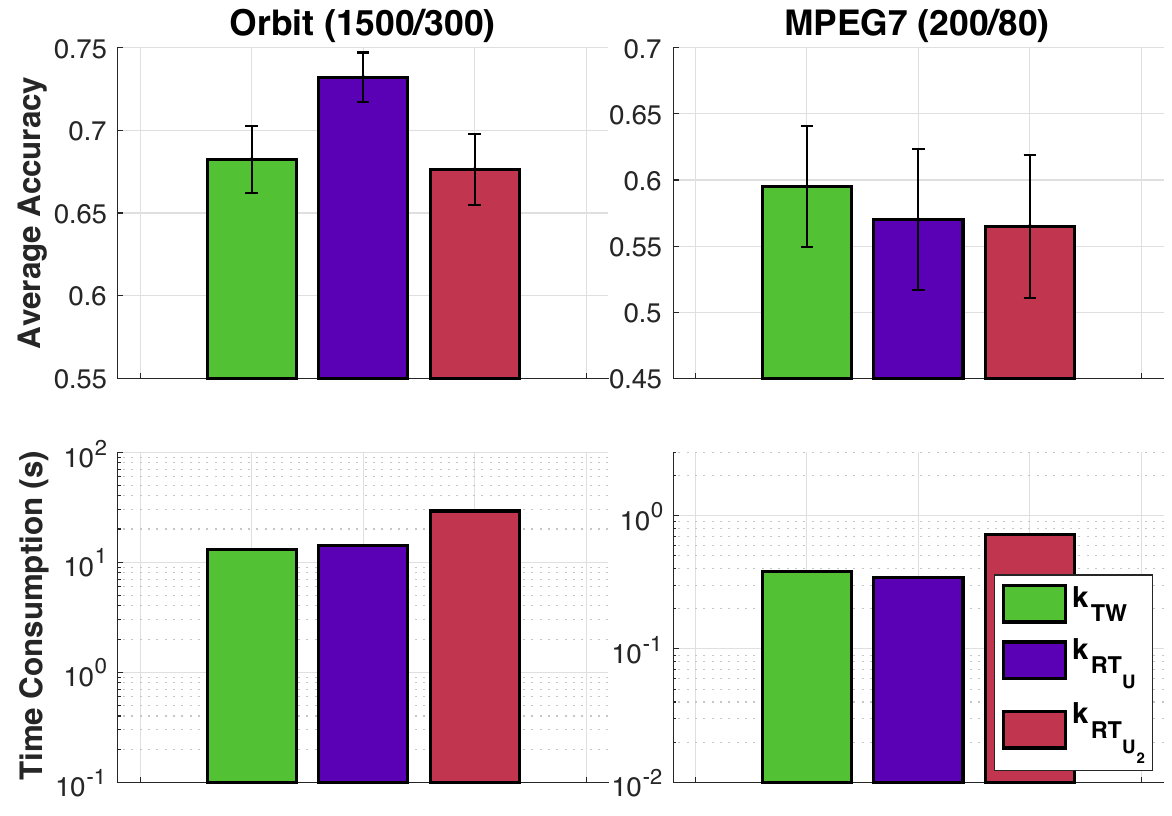}
 \end{center}
 \caption{Results on TDA for $\Delta = 0$ and $\lambda = 0.1$.}
 \label{fg:TDA_NoPerturbation_A01}
\end{figure}

\begin{figure}[h]
 \begin{center}
   \includegraphics[width=0.4\textwidth]{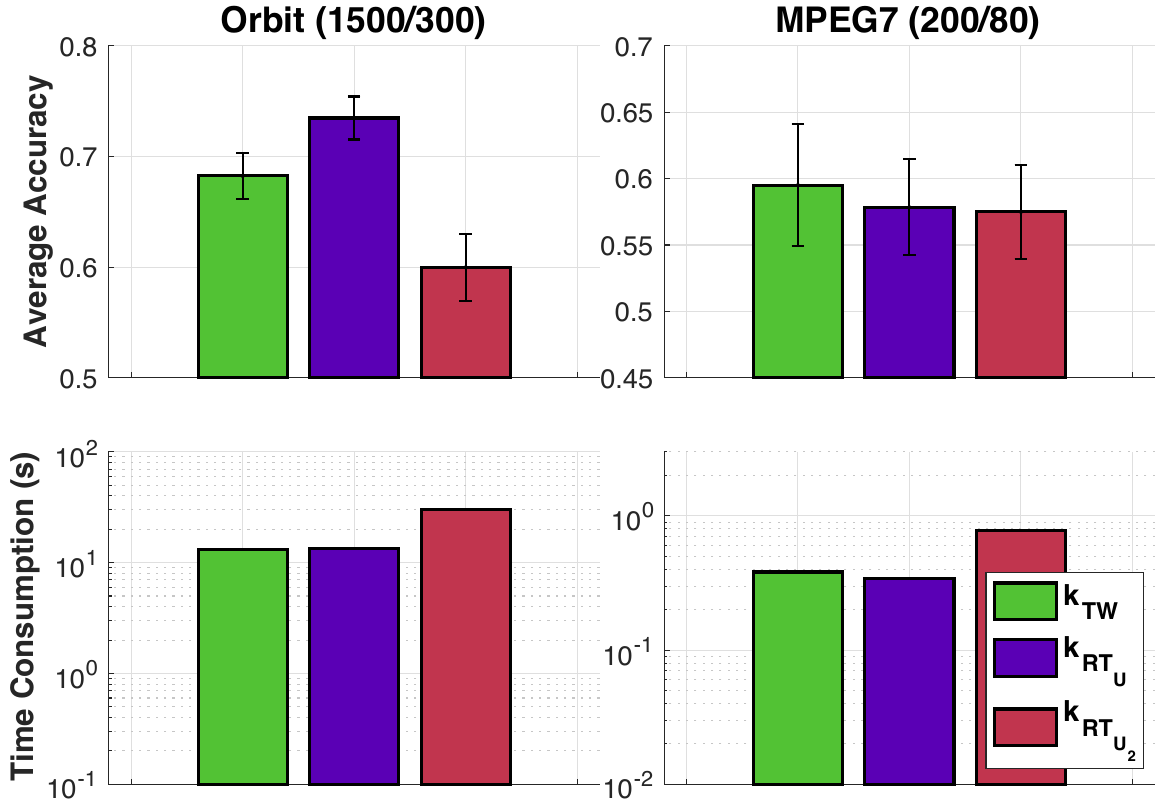}
 \end{center}
 \caption{Results on TDA for $\Delta = 0$ and $\lambda = 0.5$.}
 \label{fg:TDA_NoPerturbation_A05}
\end{figure}

\begin{figure}[h]
 \begin{center}
   \includegraphics[width=0.4\textwidth]{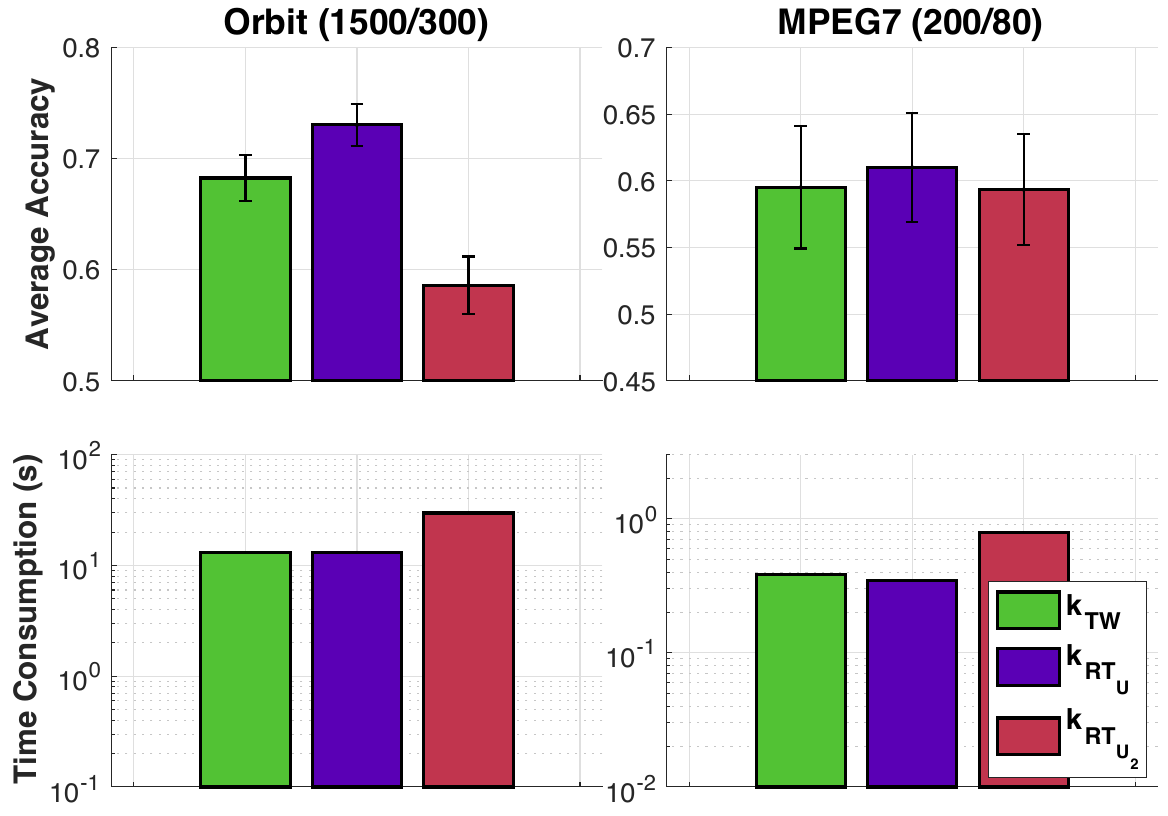}
 \end{center}
 \caption{Results on TDA for $\Delta = 0$ and $\lambda = 1$.}
 \label{fg:TDA_NoPerturbation_A1}
\end{figure}

\begin{figure}[h]
 \begin{center}
   \includegraphics[width=0.4\textwidth]{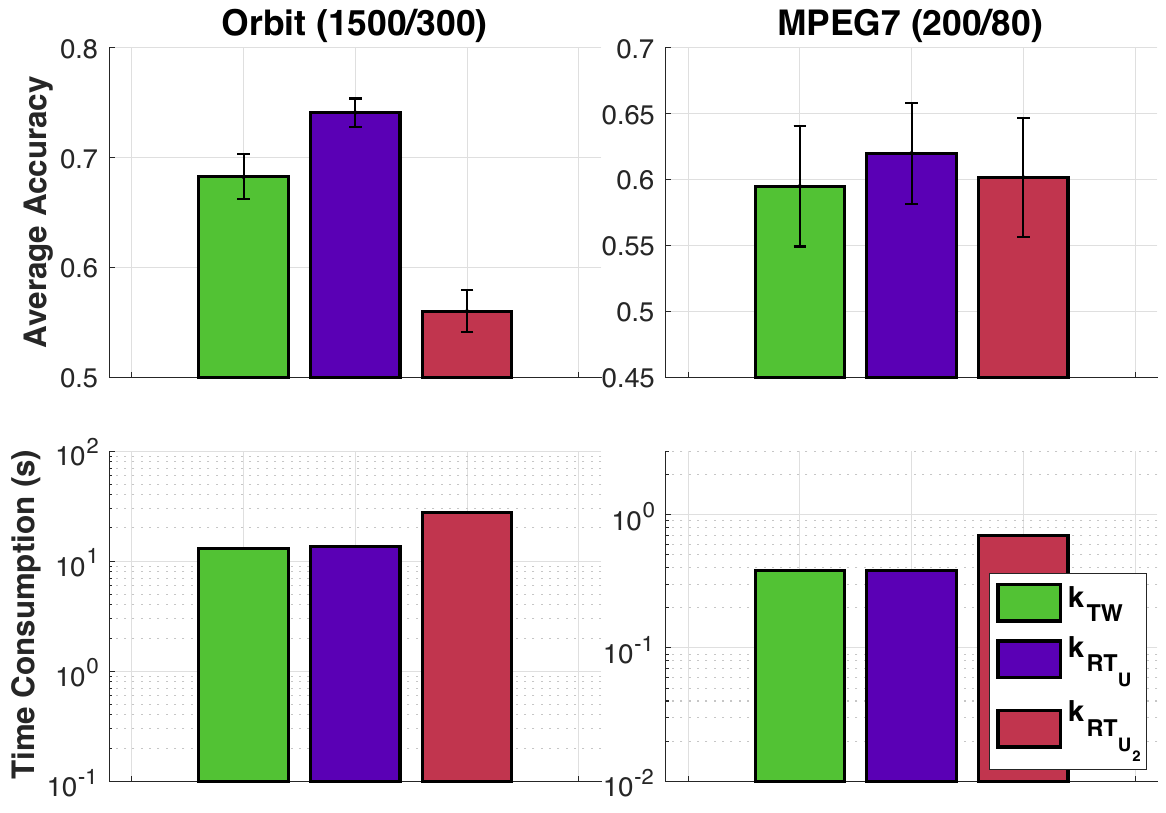}
 \end{center}
 \caption{Results on TDA for $\Delta = 0$ and $\lambda = 5$.}
 \label{fg:TDA_NoPerturbation_A5}
\end{figure}

\paragraph{For tree metric without perturbation (i.e., $\Delta = 0$).} We give detailed results for robust OT with different value of $\lambda$ in Figures~\ref{fg:TDA_NoPerturbation_A001}, \ref{fg:TDA_NoPerturbation_A005}, \ref{fg:TDA_NoPerturbation_A01}, \ref{fg:TDA_NoPerturbation_A05}, \ref{fg:TDA_NoPerturbation_A1}, and \ref{fg:TDA_NoPerturbation_A5}.

\begin{figure}[h]
 \begin{center}
   \includegraphics[width=0.4\textwidth]{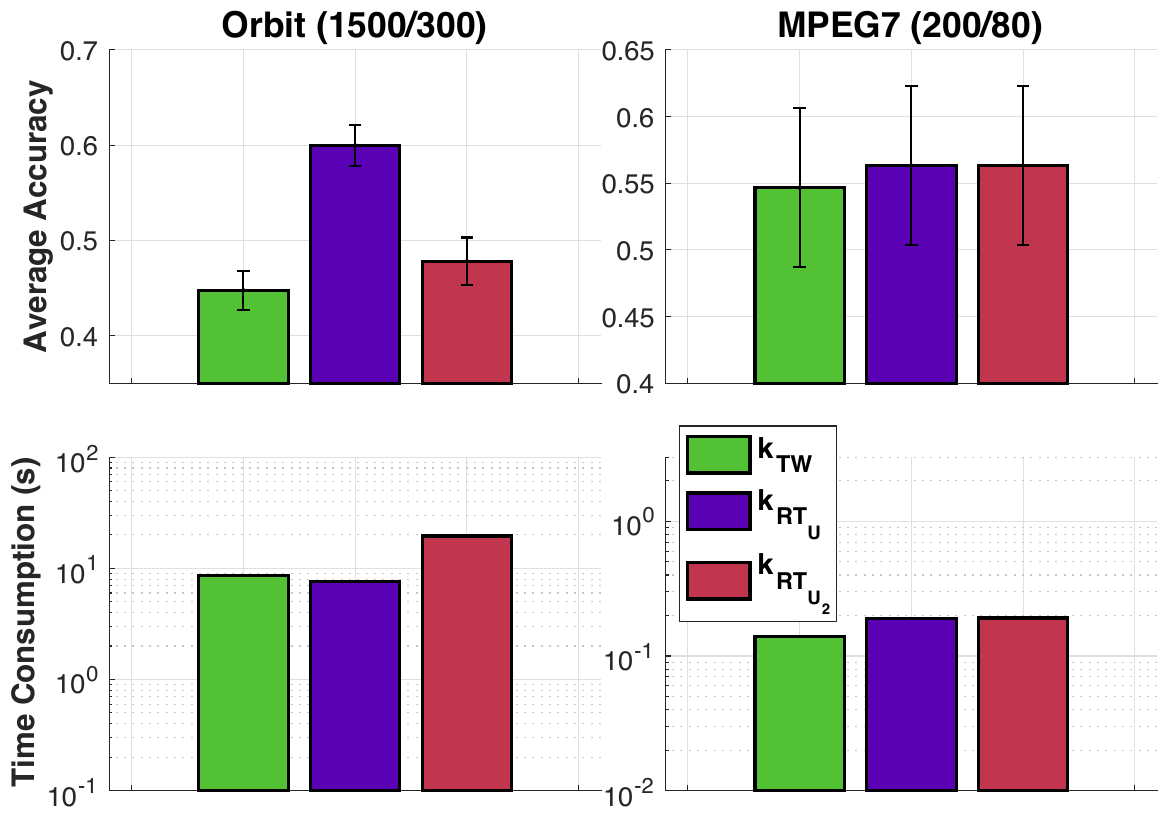}
 \end{center}
 \caption{Results on TDA for $\Delta = 0.05$ and $\lambda = 0.01$.}
 \label{fg:TDA_Noise_Delta0.05_A001}
\end{figure}

\begin{figure}[h]
 \begin{center}
   \includegraphics[width=0.4\textwidth]{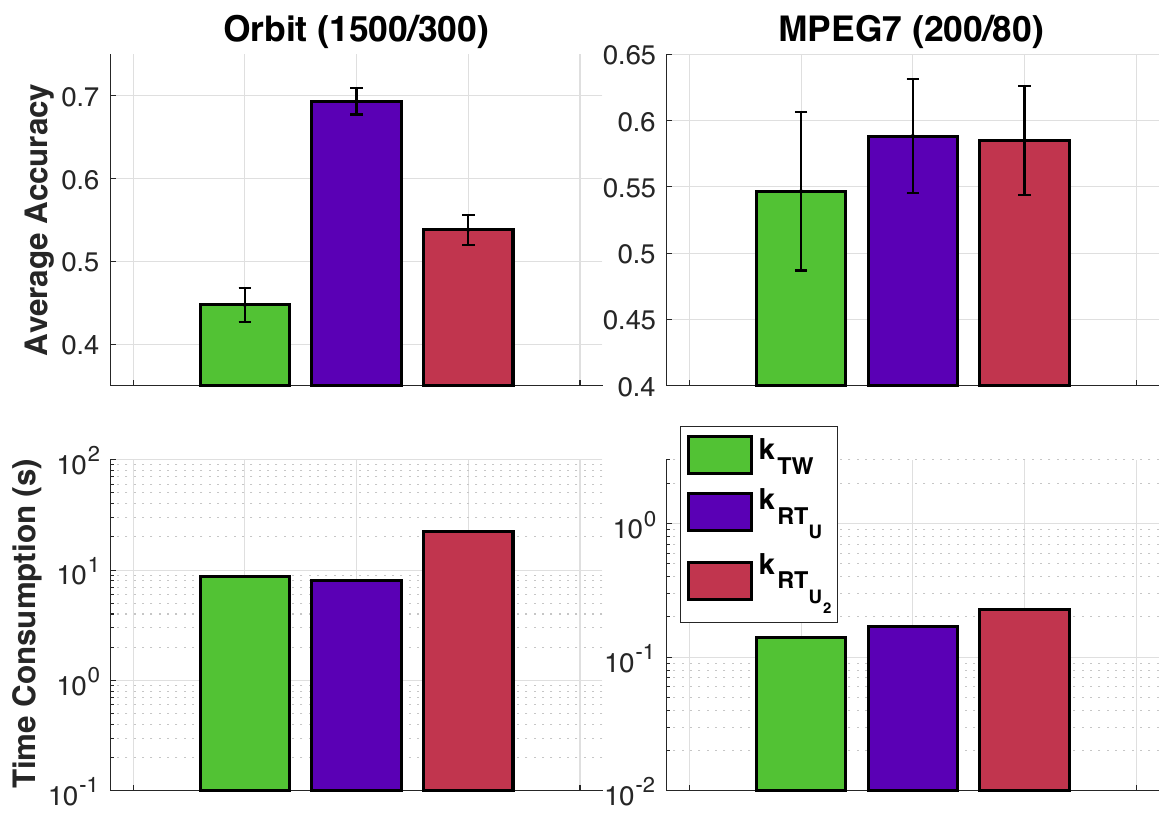}
 \end{center}
 \caption{Results on TDA for $\Delta = 0.05$ and $\lambda = 0.05$.}
 \label{fg:TDA_Noise_Delta0.05_A005}
\end{figure}

\begin{figure}[h]
 \begin{center}
   \includegraphics[width=0.4\textwidth]{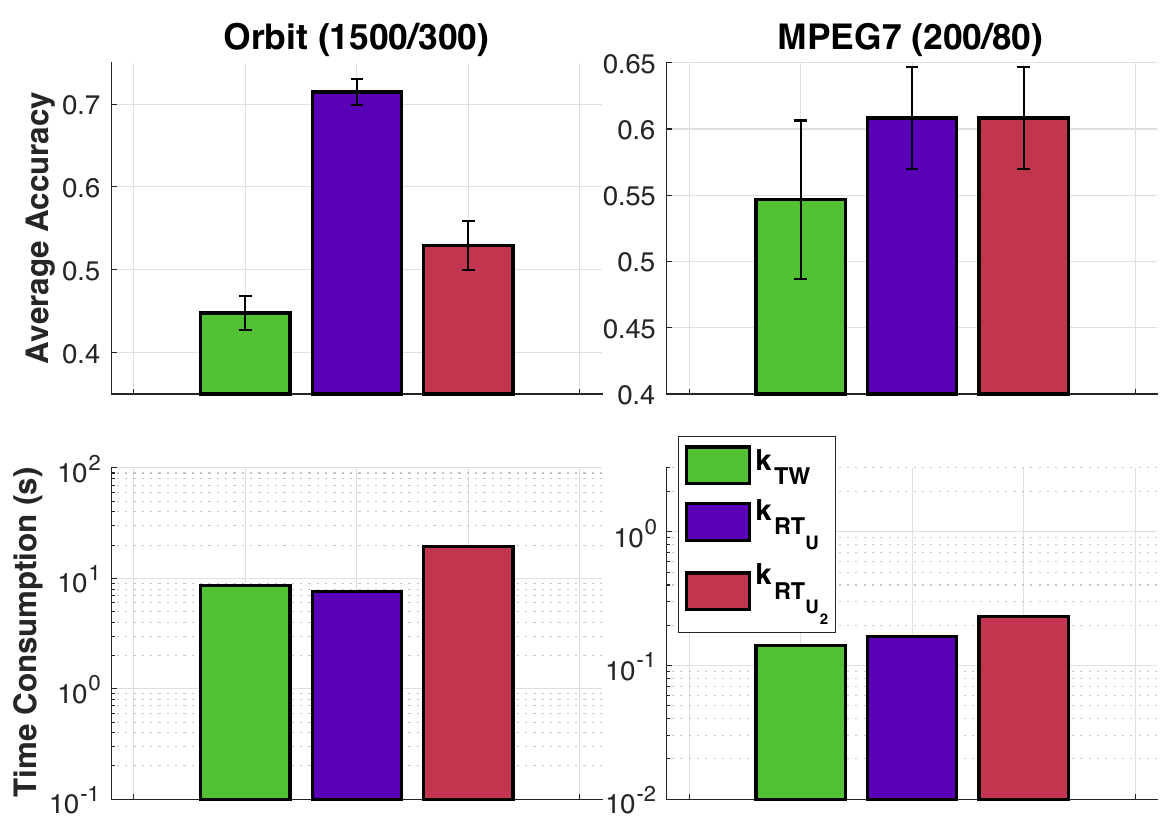}
 \end{center}
 \caption{Results on TDA for $\Delta = 0.05$ and $\lambda = 0.1$.}
 \label{fg:TDA_Noise_Delta0.05_A01}
\end{figure}

\begin{figure}[h]
 \begin{center}
   \includegraphics[width=0.4\textwidth]{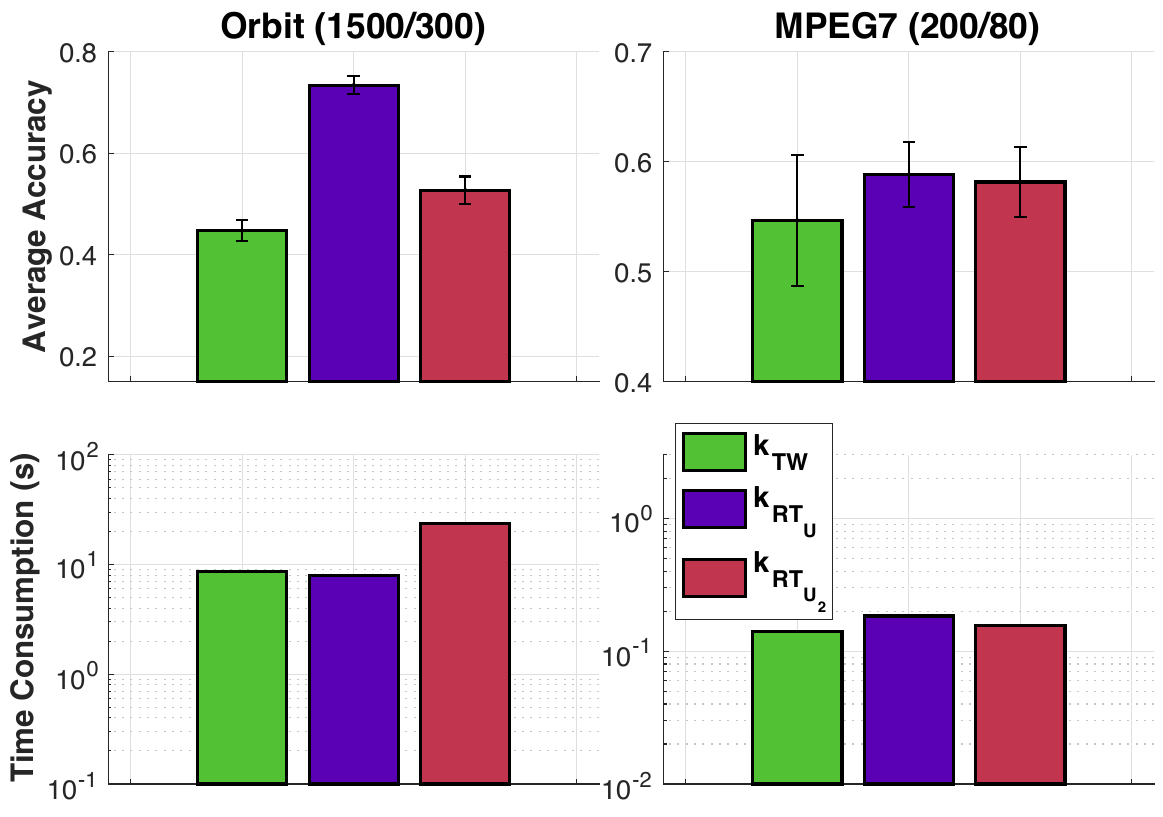}
 \end{center}
 \caption{Results on TDA for $\Delta = 0.05$ and $\lambda = 0.5$.}
 \label{fg:TDA_Noise_Delta0.05_A05}
\end{figure}

\begin{figure}[h]
 \begin{center}
   \includegraphics[width=0.4\textwidth]{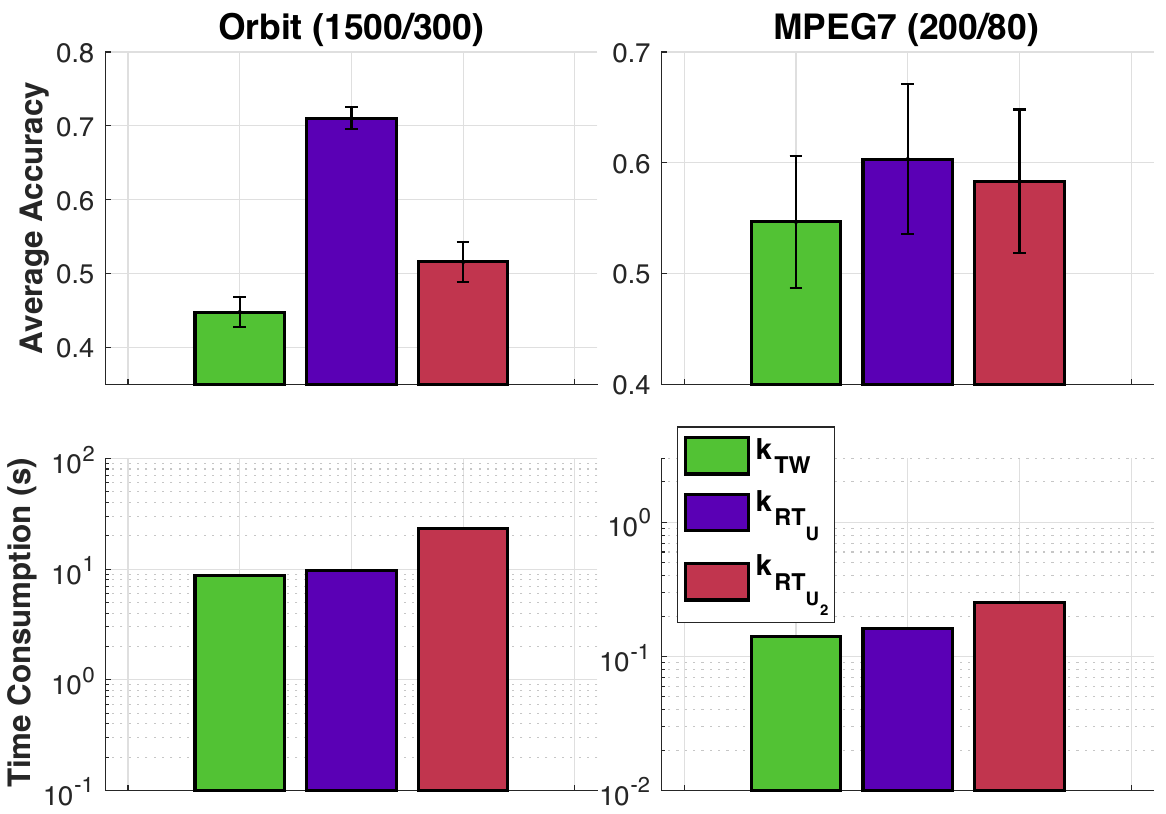}
 \end{center}
 \caption{Results on TDA for $\Delta = 0.05$ and $\lambda = 1$.}
 \label{fg:TDA_Noise_Delta0.05_A1}
\end{figure}

\begin{figure}[h]
 \begin{center}
   \includegraphics[width=0.4\textwidth]{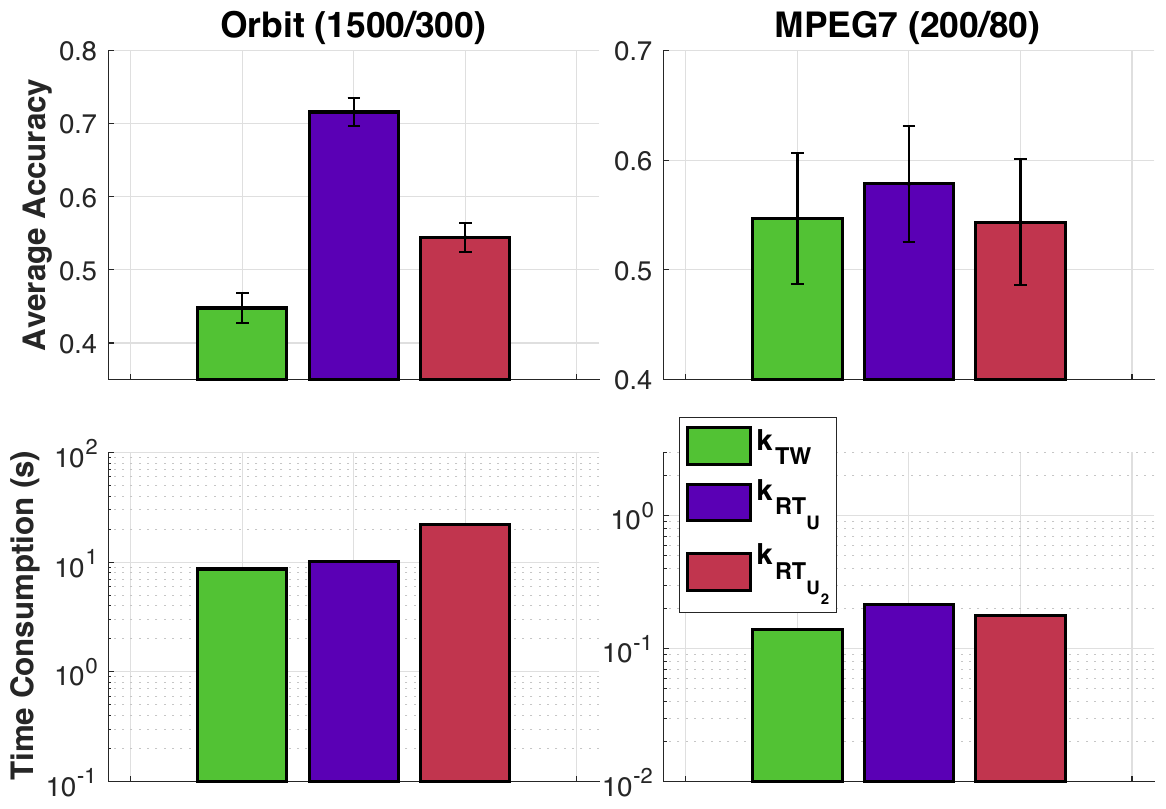}
 \end{center}
 \caption{Results on TDA for $\Delta = 0.05$ and $\lambda = 5$.}
 \label{fg:TDA_Noise_Delta0.05_A5}
\end{figure}

\paragraph{For noisy tree metric.} We give detailed results for robust OT with different value of $\lambda$ in Figures~\ref{fg:TDA_Noise_Delta0.05_A001}, \ref{fg:TDA_Noise_Delta0.05_A005}, \ref{fg:TDA_Noise_Delta0.05_A01}, 
\ref{fg:TDA_Noise_Delta0.05_A05}, \ref{fg:TDA_Noise_Delta0.05_A1}, and \ref{fg:TDA_Noise_Delta0.05_A5} when tree metric is perturbed with $\Delta = 0.05$.

%

\subsection{Discussion}

Similar to empirical results in the main manuscript, the max-min robust OT for measures with noisy tree metric is fast for computation. Their time consumptions are comparable even to that of the TW (i.e., OT with tree metric ground cost). The max-min robust OT approach helps to mitigate the issue which the given tree metric is perturbed due to noisy or adversarial measurements for OT problem. Hyperparameter $\lambda$ plays an important role for the max-min robust OT for measures with tree metric (e.g., typically chosen via cross-validation). 

\begin{table}[]
\caption{The number of pairs which we compute the distances for both training and test with kernel SVM.}
\label{tb:numpairs}
    \centering
\begin{tabular}{|l|c|}
\hline
Datasets & \#pairs \\ \hline
\texttt{TWITTER}  & 4394432 \\ \hline
\texttt{RECIPE}   & 8687560      \\ \hline
\texttt{CLASSIC}  & 22890777       \\ \hline
\texttt{AMAZON}   & 29117200      \\ \hline
\texttt{Orbit}    & 1023225   \\ \hline
\texttt{MPEG7}    & 18130     \\ \hline
\end{tabular}
\end{table}

\subsection{Further Experiments}\label{appsubsec:further_experiments}

Let consider the \texttt{TWITTER} dataset, there are $N = 3108$ documents represented as probability measures. Recall that, we randomly split $70\%/30\%$ for training and test with $10$ repeats in our experiments. Thus, for \texttt{TWITTER} dataset, the training set has $N_{tr} = 2176$ samples, and the test set has $N_{te} = 932$ samples. For the kernel SVM training, the number of pairs which we compute the distances is $(N_{tr}-1) \times \frac{N_{tr}}{2} = 2366400$. For the test phase, the number of pairs which we compute the distances is $N_{tr} \times N_{te} = 2028032$. Therefore, for $1$ repeat, the number of pairs which we compute the distances for both training and test is totally $4394432$.

When each document in \texttt{TWITTER} dataset is represented by a probability measure supported in the Euclidean space $\RR^{300}$, we randomly select $100$ pairs of probability measures to compute the subspace robust Wasserstein (SRW)~\citep{pmlr-v97-paty19a} where the dimension of the subspaces is at most $k_{SRW}=2$. The time consumption of the SRW for each pair is averagely $75.4$ seconds. Therefore, for $1$ repeat, we interpolate that the time consumption for computing the SRW for all pairs in training and test on \texttt{TWITTER} dataset should take about $4$ days averagely, while our robust TW only takes less than 100 seconds for $\RT_{\calU}$, and less than 200 seconds for $\RT_{\calU_2}$.\footnote{It takes too much time to evaluate subspace robust Wasserstein (SRW) for our experiments. Therefore, we only report the interpolation for time consumption on \texttt{TWITTER} dataset.} The time consumption issue becomes more severe on larger datasets, e.g., \texttt{AMAZON} (with more than $29$M pairs) or \texttt{CLASSIC} (with about $23$M pairs). We summarize the number of pairs which we compute their distances for both training and test for kernel SVM on all datasets in Table~\ref{tb:numpairs}.

\begin{figure}[h]
 \begin{center}
   \includegraphics[width=0.4\textwidth]{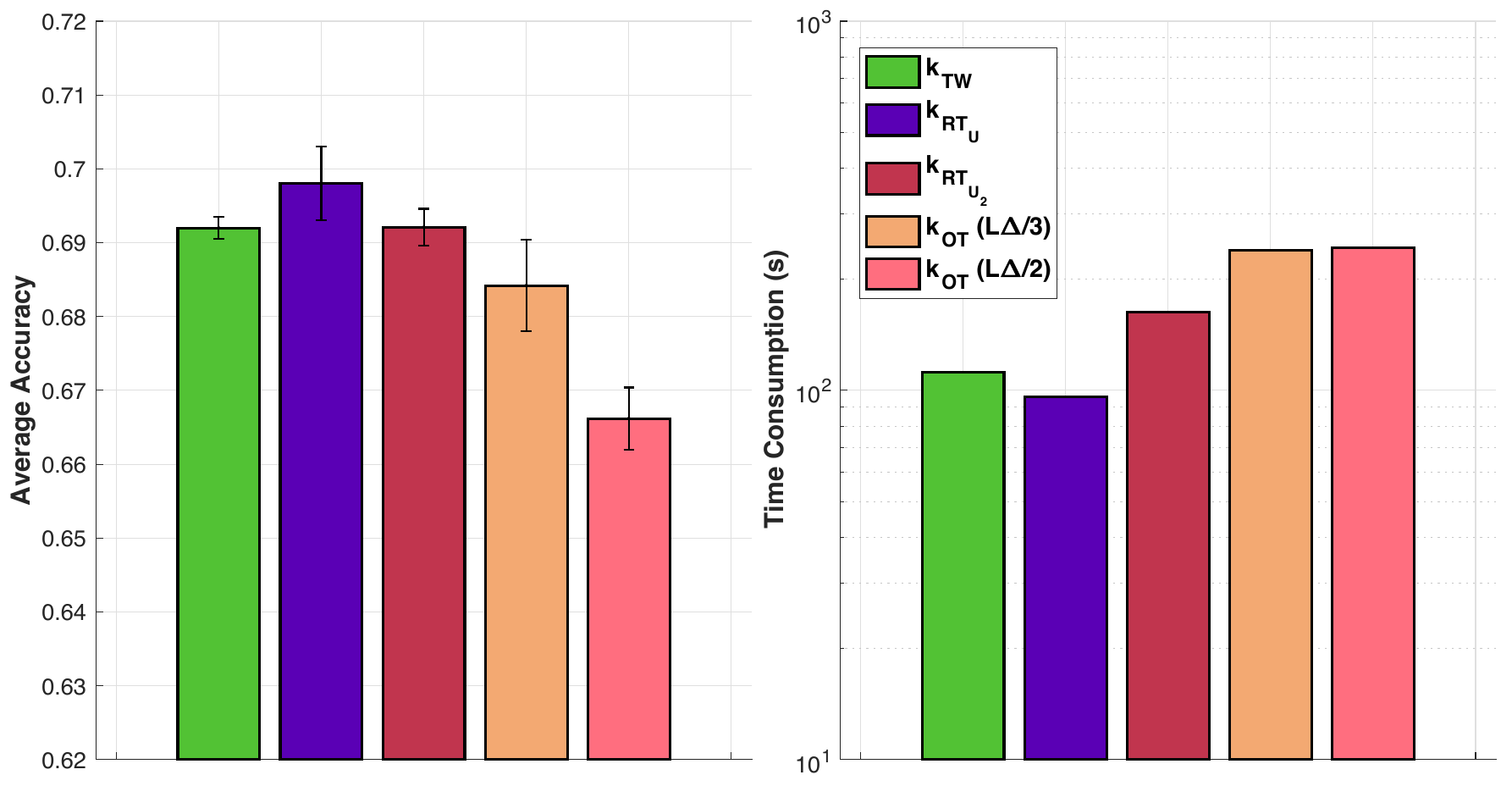}
 \end{center}
 \caption{Extended SVM results on \texttt{TWITTER} dataset.}
 \label{fg:TWITTER_Extended}
\end{figure}

For a bigger picture of empirical results, we extend the SVM results on \texttt{TWITTER} dataset in Figure~\ref{fg:DOC_Noise_Delta0.5} by adding the SVM results of the corresponding kernel for standard OT with squared Euclidean distance when each document in \texttt{TWITTER} dataset is represented as a measure supported in a high-dimensional Euclidean space $\RR^{300}$. We consider two noise levels for the ground metric with squared Euclidean distance: $\frac{L}{2}\Delta$ and $\frac{L}{3}\Delta$ where $L$ is the height of the corresponding tree metric used in the experiments for Figure~\ref{fg:DOC_Noise_Delta0.5}, and we denote them as $k_{\text{OT}} (L\Delta/3)$ and $k_{\text{OT}} (L\Delta/2)$ respectively. As noted in~\citep{LYFC}, the standard OT with squared Euclidean ground metric is indefinite and its corresponding kernel is also indefinite. We follow~\citep{LYFC} to add a sufficient regularization on its kernel Gram matrices. Figure~\ref{fg:TWITTER_Extended} illustrates this extended SVM results on \texttt{TWITTER} dataset. Although there are some differences on the experimental setting for $k_{\TW}$, $k_{\RT_{\calU}}$, $k_{\RT_{\calU_2}}$ with $k_{\text{OT}} (L\Delta/3)$, $k_{\text{OT}} (L\Delta/2)$, Figure~\ref{fg:TWITTER_Extended} illustrates an extended picture for empirical results. The performance of $k_{\text{OT}}$ is improved when the noise level is lower (i.e., $L\Delta/3$), and the indefiniteness may affect the performances of $k_{\text{OT}}$ at some certain. The time consumption of $k_{\text{OT}}$ is slower than other approaches since the time complexity of standard OT with squared Euclidean ground metric is super cubic. Our results also agree with empirical observations in~\citep{LYFC}.

%

\end{document}